\documentclass{article}
\usepackage[T1]{fontenc}
\usepackage[utf8]{inputenc}
\usepackage{array}
\usepackage{float}
\usepackage{multirow}
\usepackage{amsmath}
\usepackage{amssymb}
\usepackage{stmaryrd}
\usepackage{graphicx}
\usepackage[authoryear]{natbib}
\usepackage{microtype}
\usepackage[unicode=true,
 bookmarks=false,
 breaklinks=false,pdfborder={0 0 1},backref=section,colorlinks=false]
 {hyperref}
\hypersetup{pdftitle={A template for the arxiv style},
 pdfauthor={David S.~Hippocampus, Elias D.~Striatum},
 pdfsubject={q-bio.NC, q-bio.QM},
 pdfkeywords={First keyword, Second keyword, More}}

\makeatletter

\providecommand{\tabularnewline}{\\}

\usepackage{arxiv}

\usepackage{url}% simple URL typesetting
\usepackage{amsfonts}% blackboard math symbols
\usepackage{nicefrac}% compact symbols for 1/2, etc.
\usepackage{lipsum}% Can be removed after putting your text content
\usepackage{graphicx}
\usepackage{doi}

\usepackage{float} 
\usepackage{graphicx}
\usepackage{microtype}	
\usepackage{tabularx}

\usepackage{tikz}
\usepackage{comment}
\usepackage{color}

\title{A template for the \emph{arxiv} style}

\author{%
  Thao Minh Le, Vuong Le, Sunil Gupta, Svetha Venkatesh, Truyen Tran \\
  Applied Artificial Intelligence Institute, Deakin University, Australia\\
  \texttt{\{lethao,vuong.le,sunil.gupta,svetha.venkatesh,truyen.tran\}@deakin.edu.au} \\
}

\date{}

\makeatother

\begin{document}
\title{Guiding Visual Question Answering with Attention Priors}

\maketitle
\global\long\def\Problem{\text{VQA}}%

\global\long\def\eg{e.g.}%

\begin{abstract}
The current success of modern visual reasoning systems is arguably
attributed to cross-modality attention mechanisms. However, in deliberative
reasoning such as in VQA, attention is unconstrained at each step,
and thus may serve as a statistical pooling mechanism rather than
a semantic operation intended to select information relevant to inference.
This is because at training time, attention is only guided by a very
sparse signal (i.e. the answer label) at the end of the inference
chain. This causes the cross-modality attention weights to deviate
from the desired visual-language bindings. To rectify this deviation,
we propose to guide the attention mechanism using explicit linguistic-visual
grounding. This grounding is derived by connecting structured linguistic
concepts in the query to their referents among the visual objects.
Here we learn the grounding from the pairing of questions and images
alone, without the need for answer annotation or external grounding
supervision. This grounding guides the attention mechanism inside
VQA models through a duality of mechanisms: pre-training attention
weight calculation and directly guiding the weights at inference time
on a case-by-case basis. The resultant algorithm is capable of probing
attention-based reasoning models, injecting relevant associative knowledge,
and regulating the core reasoning process. This scalable enhancement
improves the performance of VQA models, fortifies their robustness
to limited access to supervised data, and increases interpretability.

\end{abstract}

\section{Introduction}

Visual reasoning is the new frontier of AI wherein facts extracted
from visual data are gathered and distilled into higher-level knowledge
in response to a query. Successful visual reasoning methodology estimates
the cross-domain association between the symbolic concepts and visual
entities in the form of attention weights. Such associations shape
the knowledge distillation process, resulting in a unified representation
that can be decoded into an answer. In the exemplar reasoning setting
known as Visual Question Answering (VQA), attention plays a pivotal
role in modern systems \citep{anderson2018bottom,hudson2018compositional,kim2018bilinear,le2020dynamic,lu2016hierarchical}.
Ideal attention scores must be both \emph{relevant} and \emph{effective}:
Relevance implies that attention is high when the visual entity and
linguistic entity refer to the same concept; Effectiveness implies
that the attention derived leads to good VQA performance.

However, in typical systems, the attention scores are computed on-the-fly:
unregulated at inference time and guided at training time by the gradient
from the groundtruth answers. Analysis of several VQA attention models
shows that these attention scores are usually neither relevant nor
guaranteed to be effective \citep{das2017human}. The problem is even
more severe when we cannot afford to have enough labeled answers due
to the cost of the human annotation process. A promising solution
is providing pre-computed guidance to direct and hint the attention
mechanisms inside the VQA models towards more appropriate scores.
Early works use human attention as the label for supervising machine
attention \citep{qiao2018exploring,selvaraju2019taking}. This simple
and direct attention perceived by humans is not guaranteed to be optimal
for machine reasoning \citep{firestone2020performance,fleuret2011comparing}.
Furthermore, because annotating attention is a complex labeling task,
this process is inherently costly, inconsistent and unreliable \citep{selvaraju2019taking}.
Finally, these methods only regulate the attention scores in training
stage without directly adjust them in inference. Different from these
approaches, we leverage the fact that such external guidance is pre-existing
in the query-image pairs and can be extracted without any additional
labels. \emph{Using pre-computed language-visual associations as an
inductive bias for attention-based reasoning without further extra
labeling remains a desired but missing capability.}

\begin{figure}[t]
\begin{centering}
\includegraphics[width=0.6\columnwidth]{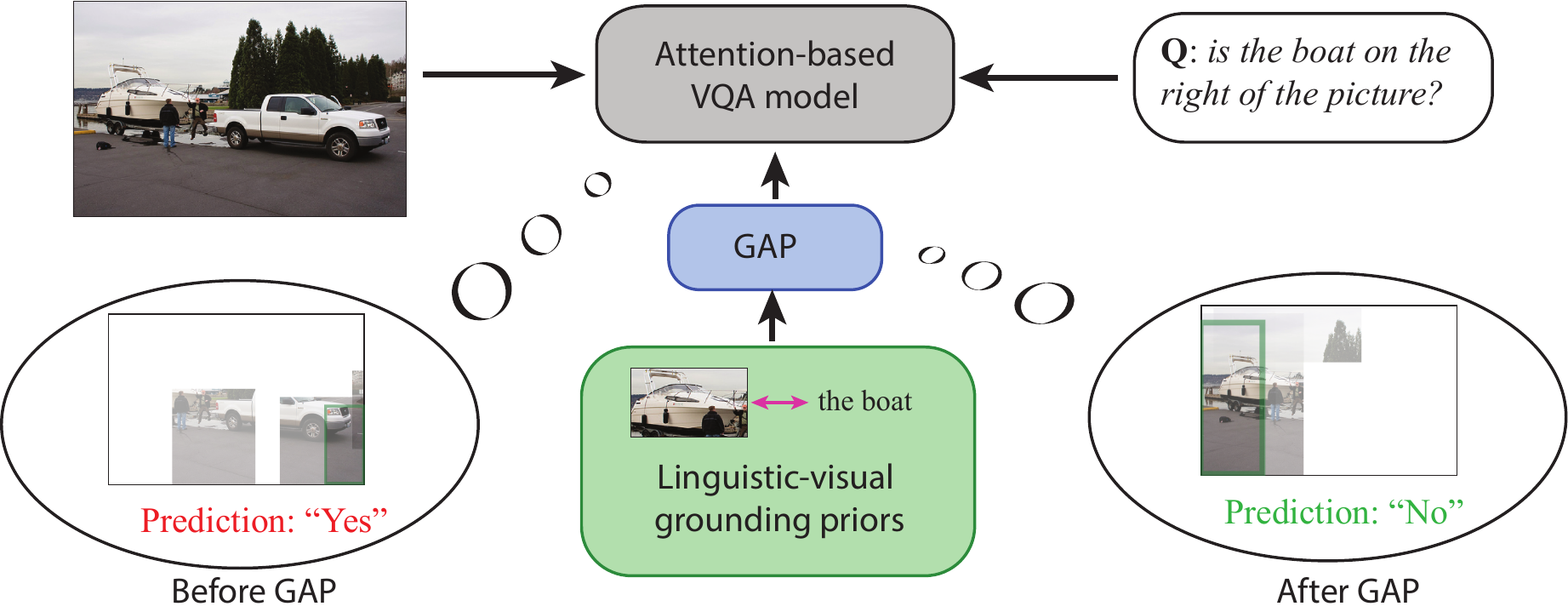}
\par\end{centering}
\caption{We introduce Grounding-based Attention Prior (GAP) mechanism (blue
box) which considers the linguistic-visual associations between a
pair of VQA query and image and refines the attentions inside reasoning
model (gray box). This boosts the performance of VQA models, reduces
their reliance on supervised data and increases their interpretability.
\label{fig:teaser}}
\end{figure}

Exploring this underlying linguistic-visual association for VQA, we
aim to distill the compatibility between entities across input modalities
in an unsupervised manner from the query-image pairs without explicit
alignment grouthtruths, and use this knowledge as an inductive bias
for the attention mechanism thus boosting reasoning capability. To
this end, we design a framework called \emph{Grounding-based Attention
Prior }(GAP) to (1) extract the alignments between linguistic-visual
region pairs and (2) use these pair-wise associations as an inductive
bias to guide VQA's attention mechanisms.

For the first task, we exploit the pairing between the questions and
the images as a weakly supervised signal to learn the mapping between
words and image regions. By exploiting the implicit supervising signals
from the pairing, this requires no further annotation. To overcome
the challenge of disparity in the co-inferred semantics between query
words and image regions, we construct a parse tree of the query, extract
the nested phrasal expressions and ground them to image regions. These
expressions semantically match image regions better than single words
and thus create a set of more reliable linguistic-visual alignments.

The second task aims at using these newly discovered alignments to
guide reasoning attention. This guidance process is provided through
two complementary pathways. First, we pre-train attention weights
to align with the pre-computed grounding. This step is done in an
unsupervised manner without access to the answer groundtruths. Second,
we use the attention prior to directly regulate and refine the attention
weights guided by the groundtruth answer through back-propagation
to not deviate too far away from it. This is modulated by a learnable
gate. These dual guidance pathways are a major advancement from previous
attention regularization methods \citep{selvaraju2019taking,wu2019self}
as the linguistic-visual compatibility is leveraged directly and flexibly
in both training and inference rather than simply as just regularization.

Through extensive experiments, we prove that this methodology is effective
in both discovering the grounding and using them to boost the performance
of attention-based VQA models across representative methods and datasets.
These improvements surpass other methods' performance and furthermore
require no extra annotation. The proposed method also significantly
improves the sample efficiency of VQA models, hence less annotated
answers are required. Fig.~\ref{fig:teaser} illustrates the intuition
and design of the method with an example of the improved attention
and answer.

Our key contributions are:

1. A novel framework to calculate linguistic-visual alignments, providing
pre-computed attention priors to guide attention-based VQA models;

2. A generic technique to incorporate attention priors into most common
visual reasoning methods, fortifying them in performance and significantly
reducing their reliance on human supervision; and,

3. Rigorous experiments and analysis on the relevance of linguistic-visual
alignments to reasoning attention.

\section{Related Work \label{sec:Related-Work}}

\textbf{Attention-based models} are the most prominent approaches
in VQA. Simple methods \citep{anderson2018bottom} only used single-hop
attention mechanism to help machine select relevant image features.
More advanced methods \citep{yang2016stacked,hudson2018compositional,le2020dynamic}
and those relying on memory networks \citep{xiong2016dynamic,xu2016ask}
used multi-hop attention mechanisms to repeatedly revise the selection
of relevant visual information. BAN \citep{kim2018bilinear} learned
a co-attention map using expensive bilinear networks to represent
the interactions between pairs of word-region. One drawback of these
attention models is that they are only supervised by the answer groundtruth
without explicit attention supervision. 

\textbf{Attention supervision }is recently studied for several problems
such as machine translation \citep{liu2016neural} and image captioning
\citep{liu2017attention,ma2020learning,zhou2020more}. In VQA, attentions
can be self-regulated through internal constraints \citep{ramakrishnan2018overcoming,liu2021answer}.
More successful regularization methods use external knowledge such
as human annotations on textual explanations \citep{wu2019self} or
visual attention \citep{qiao2018exploring,selvaraju2019taking}. Unlike
these, we propose to supervise VQA attentions using pre-computed
language-visual grounding from image-query pairs without using external
annotation. 

\textbf{Linguistic-visual alignment} includes the tasks of text-image
matching \citep{lee2018stacked}, grounding referring expressions
\citep{yu2018mattnet} and cross-domain joint representation \citep{lu2019vilbert,su2020vl}.
These groundings can support tasks such as captioning \citep{zhou2020more,karpathy2015deep}.
Although most tasks are supervised by human annotations, contrastive
learning \citep{gupta2020contrastive,wang2021improving} allows machines
to learn the associations between words and image regions from weak
supervision of phrase-image pairs. In this work, we propose to explore
such associations between query and image in VQA. This is a new challenge
because the query is complex and harder to be grounded, therefore
new method using grammatical structure will be devised. 

Our work also share the \textbf{Knowledge distillation} paradigm \citep{hinton2015distilling}
with cross-task \citep{albanie2018emotion} and cross modality \citep{gupta2016cross,liu2018multi,wang2020improving}
adaptations. Particularly, we distill visual-linguistic grounding
and use it as an input for VQA model's attention. This also distinguishes
our work from the recent \textbf{self-supervised pretraining methods}
\citep{tan2019lxmert,li2020oscar} where they focus on a unified representation
for a wide variety of tasks thanks to the access to enormous amount
of data. Our work is theoritically applicable to complement the multimodal
matching inside these models.

\section{Preliminaries\label{sec:Preliminary}}

A VQA system aims to deduce an answer $y$ about an image $\mathcal{I}$
in response to a linguistic question $q$, $\eg$, via $P(y\mid q,\mathcal{I})$.
The query $q$ is typically decomposed into a set of $T$ linguistic
entities $L=\left\{ l_{i}\right\} _{i=1}^{T}$. These entities and
the query $q$ are then embedded into a feature vector space: $q\in\mathbb{R}^{d},\,l_{i}\in\mathbb{R}^{d}$.
In the case of sequential embedding popularly used for VQA, entities
are query words; they are encoded with GloVe for word-level embedding
\citep{pennington2014glove} followed by RNNs such as BiLSTM for sentence-level
embedding. Likewise the image $\mathcal{I}$ is often segmented into
a set of $N$ visual regions with features $V=\left\{ v_{j}\mid v_{j}\in\mathbb{R}^{d}\right\} _{j=1}^{N}$
by an object detector, i.e., Faster R-CNN \citep{ren2015faster}.
For ease of reading, we use the dimension $d$ for both linguistic
embedding vectors and visual representation vectors.

A large family of VQA systems \citep{lu2016hierarchical,anderson2018bottom,hudson2018compositional,le2020dynamic,kim2018bilinear,kim2016hadamard}
rely on attention mechanisms to distribute conditional computations
on linguistic entities $L$ and visual counterparts $V$. These models
can be broadly classified into two groups: \emph{joint-} and \emph{marginalized-
attention models}. \citet{lu2016hierarchical,anderson2018bottom,hudson2018compositional}
are among those who fall into the former, while \citet{kim2018bilinear,kim2016hadamard}
and Transformer-based models \citep{tan2019lxmert} are typical representative
of the works in the latter category.

\paragraph{Joint attention models}

The most complete attention model includes a detailed pair-wise attention
map indicating the contextualized correlation between word-region
pairs used to estimate the interaction between visual and linguistic
entities for the combined information. These attention weights are
in the form of a 2D matrix $A\in\mathbb{R}^{T\times N}$. They often
contain fine-grained relationships between each linguistic word to
each visual region. The attention matrix $A$ is derived by a sub-network
$B_{\theta}(.)$ as $A_{ij}=B_{\theta}(e_{ij}\mid V,L)$, where each
$e_{ij}$ denotes the correlation between the linguistic entities
$l_{i}$ and the visual region $v_{j}$, and $\theta$ is network
parameters of VQA models. Joint attention models contain the rich
pairwise relation and often perform well. However, calculating and
using this full matrix has a large overhead computation cost. A good
approximation of this matrix is the marginalized vectors over rows
and columns which is described next.

\paragraph{Marginalized attention models}

Conceptually, the matrix $A$ is marginalized along columns into the
\emph{linguistic attention vector} $\alpha=\left\{ \alpha_{i}\right\} _{i=1}^{T}\in\mathbb{R}^{T}$
and along rows into \emph{visual attention vector} $\beta=\left\{ \beta_{j}\right\} _{j=1}^{N},\beta\in\mathbb{R}^{N}$.
 In practice, $\alpha$ and $\beta$ are calculated directly from
each pair of input image and query through dedicated attention modules.
They can be implemented in different ways such as direct single-shot
attention \citep{anderson2018bottom}, co-attention \citep{lu2016hierarchical}
or multi-step attention \citep{hudson2018compositional}. In our
experiment, we concentrate on two popular mechanisms: \emph{single-shot
attention} where the visual attention $\beta$ is calculated directly
from the inputs $(V,q)$ and \emph{alternating attention mechanism}
where the visual attention $\beta$ follows the linguistic attention
$\alpha$ \citep{lu2016hierarchical}. Concretely, $\alpha$ is estimated
first, followed by the attended linguistic feature of the entire query
$c=\sum_{i=1}^{T}\alpha_{i}*l_{i}$; then this attended linguistic
feature is used to calculate the visual attention $\beta$. The alternating
mechanism can be extended with multi-step reasoning \citep{hudson2018compositional,le2020dynamic,hu2019language}.
In such case, a pair of attentions $\alpha_{i,k}$ and $\beta_{j,k}$
are estimated at each reasoning step $k$ forming a series of them.

\begin{figure*}[t]
\begin{centering}
\includegraphics[width=1\textwidth]{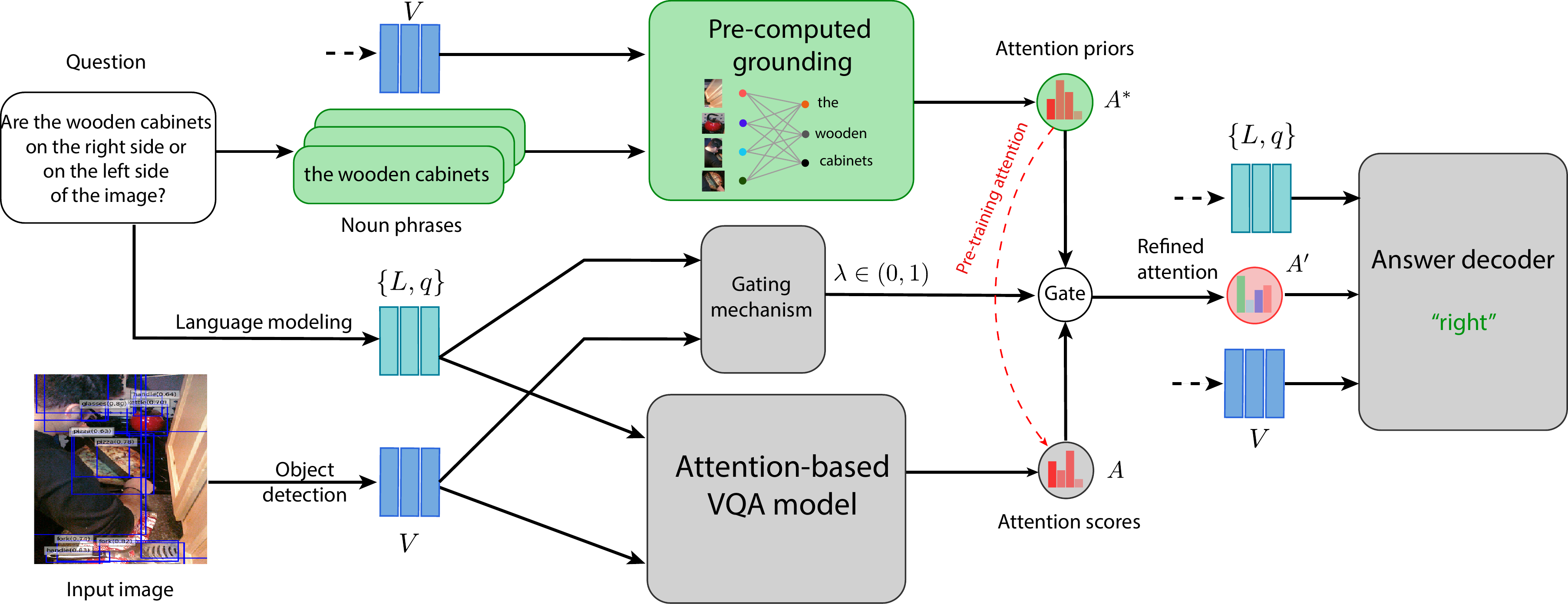}
\par\end{centering}
\caption{Overall architecture of a generic joint attention VQA model using
Grounding-based Attention Prior (GAP) to guide the computation of
attention weights. Vision-language compatibility pre-computed by an
unsupervised framework (green boxes) serves as an extra source of
information, providing inductive biases to guide attention weights
inside attention-based VQA models towards more meaningful alignment.\label{fig:Overall}}
\end{figure*}

\paragraph{Answer decoder}

Attention scores drive the reasoning process producing a joint linguistic-visual
representation on which the answer is decoded: \sloppy$P\left(a\mid f(L,V,\text{att\_scores})\right)$
(``att\_scores'' refers to either visual attention vector $\beta$
or attention matrix $A$). For marginalized attention models, the
function $f(.)$ is a neural network taking as input the query representation
$q$ and the attended visual feature $\hat{v}=\sum_{j}\beta_{j}*v_{j}$
to return a joint representation. Joint attention models instead use
the bilinear combination to calculate each component of the output
vector of $f$ \citep{kim2018bilinear}:

\begin{equation}
f_{t}\equiv(L^{\top}W_{L})_{t}^{\top}A(V^{\top}W_{V})_{t},
\end{equation}
where $t$ is the index of output components and $W_{L}\in\mathbb{R}^{d\times d}$
and $W_{V}\in\mathbb{R}^{d\times d}$ are learnable weights.

\section{Methods\label{sec:Methods}}

We now present Grounding-based Attention Priors (GAP), an approach
to extract the concept-level association between query and image and
use this knowledge as attention priors to guide and refine the cross-modality
attentions inside VQA systems. The approach consists of two main stages.
First, we learn to estimate the linguistic-visual alignments directly
from question-image pairs (Sec.~\ref{sec:Grounding}, green boxes
in Fig.~\ref{fig:Overall}). Second, we use such knowledge as inductive
priors to assist the computation of attention in VQA (Sec.~\ref{sec:Pretraining},
Sec.~\ref{sec:refinement}, and lower parts in Fig.~\ref{fig:Overall}).

\subsection{Structures for Linguistic-Visual Alignment\label{sec:Grounding}}

\begin{figure}[t]
\begin{centering}
\includegraphics[width=0.58\columnwidth]{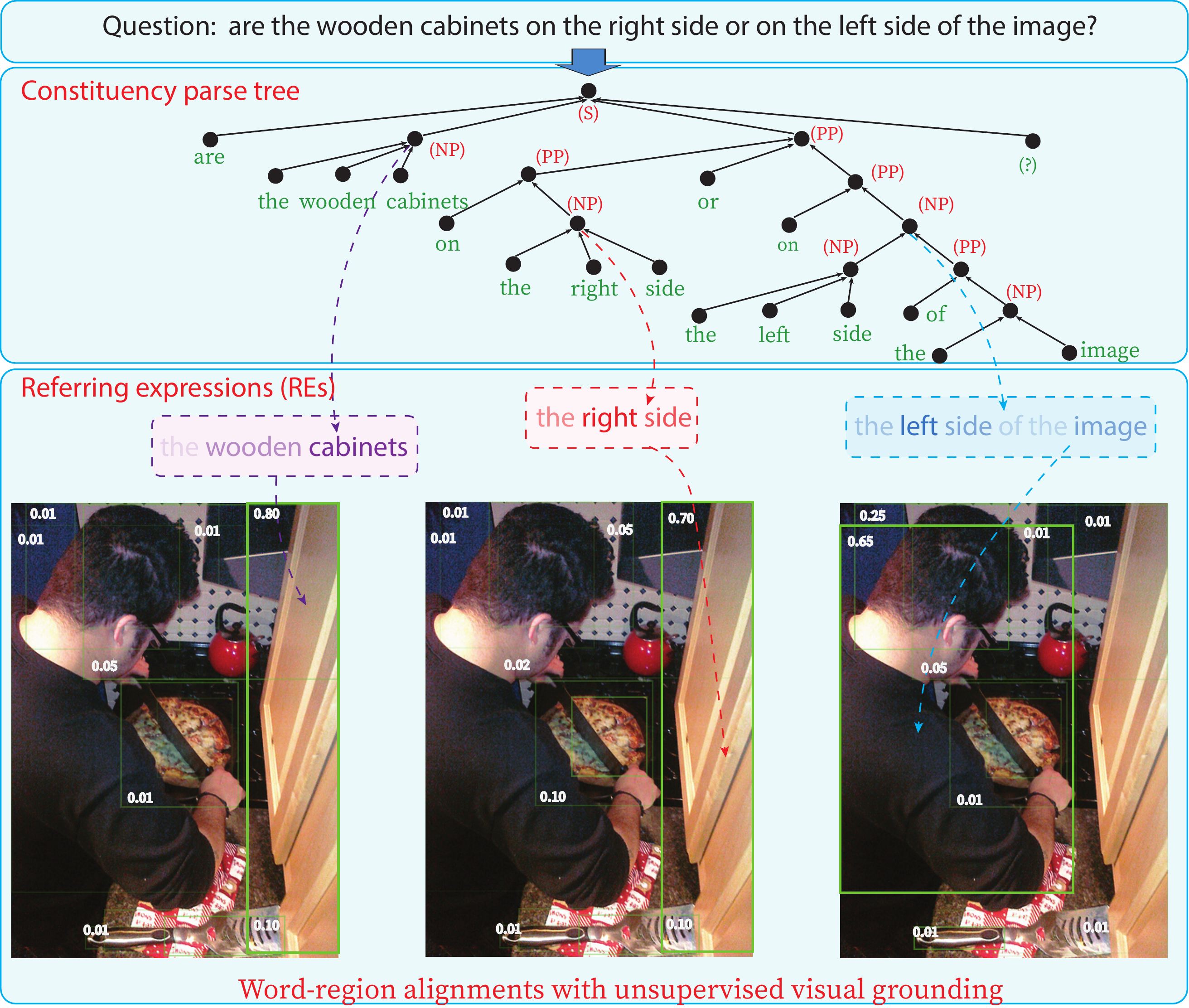}
\par\end{centering}
\caption{The query is parsed into a constituency parse tree to identify REs.
 Each RE serves as a local context for words. Words within each RE
context are grounded to corresponding image regions. A word can appear
in multiple REs, and thus its final grounding is averaged over containing
REs, serving as inductive prior for VQA.\label{fig:Groundeing_illustration}}
\end{figure}

\textbf{Grammatical structures for grounding}. The task of \emph{Linguistic-visual
Alignment} aims to find the groundings between the linguistic entities
($\eg$, query words $L=\left\{ l_{i}\right\} _{i=1}^{T}$ in VQA)
and vision entities ($\eg$, visual regions $V=\left\{ v_{j}\right\} _{j=1}^{N}$
in VQA) in a shared context. This requires the interpretation of individual
words in the complex context of the query so that they can co-refer
to the same concepts as image regions. However, compositional queries
have complex structures that prevent state-of-the-art language representation
methods from fully understanding the relations between semantic concepts
in the queries \citep{reimers2019sentence}. We propose to better
contextualize query words by \emph{breaking a full query into phrases
that refer to simpler structures, }making the computation of word-region
grounding more effective\emph{. }These phrases are called \emph{referring
expressions} (RE) \citep{mao2016generation} and were shown to co-refer
well to image regions \citep{kazemzadeh2014referitgame}. The VQA
image-query pairing labels are passed to the REs of such query. We
then ground words with contextualized embeddings within each RE to
their corresponding visual regions. As the REs are nested phrases
from the query, a word can appear in multiple REs. Thus, we obtain
the query-wide \emph{word-region grounding} by aggregating the grounding
of REs containing the word. See Fig.~\ref{fig:Groundeing_illustration}
for an example on this process.

We extract query REs using a constituency parse tree $\mathcal{T}$\citep{cirik2018using}.\footnote{Berkeley Neural Parser \citep{kitaev2018constituency} in our implementation.}
In this structure, the query is represented as a set of nested phrases
corresponding to subtrees of $\mathcal{T}$. The parser also provides
the grammatical roles of the phrases. For example,  the phrase ``the
white car'' will be tagged as a \emph{noun-phrase }while ``standing
next to the white car'' is a\emph{ verb-phrase}. As visual objects
and regions are naturally associated with noun-phrases, we select
a set $E=\left\{ E_{r}\right\} $ of all the noun phrases and wh-noun
phrases\footnote{noun phrases prefixed by a pronoun, $\eg$, ``which side'', ``whose
bag''.} as the REs. 

We denote the $r^{th}$ RE as $E_{r}\hspace{-1mm}=\hspace{-1mm}\{w_{i}\hspace{-1mm}\mid\hspace{-1mm}w_{i}\in\mathbb{R}^{d}\}_{s_{r}\leq i\leq e_{r}}$
where $s_{r}$ and $e_{r}$ are the start and end word index of the
RE within the query $L=\left\{ l_{i}\right\} _{i=1}^{T}$. It has
length $m_{r}=e_{r}-s_{r}+1$. We now estimate the correlation between
words in these REs and the visual regions $V=\left\{ v_{j}\mid v_{j}\in\mathbb{R}^{d}\right\} _{j=1}^{N}$
by learning the neural association function $g_{\delta}(V,E_{r})$
of parameter $\delta$ that generates a mapping $A_{r}^{*}\in\mathbb{R}^{m_{r}\times N}$
between words in the RE and the corresponding visual regions. 

We implement $g_{\delta}\left(.\right)$ as the dot products of a
contextualized embedding of word $w_{i}$ in $E_{r}$ with the representations
of regions in $V$, following the scaled dot-product attention \citep{vaswani2017attention}.

\textbf{Unsupervised training.} To train the function $g_{\delta}\left(.\right)$,
we adapt the recent contrastive learning framework \citep{gupta2020contrastive}
for phrase grounding to learn these word-region alignments from the
RE-image pairs in an unsupervised manner, i.e. without explicit word-region
annotations. In a mini batch $\mathcal{B}$ of size $b$, we calculate
the positive mapping $A_{r}^{*}=\left(a_{r,i,j}^{*}\right)\in\mathbb{R}^{m_{r}\times N}$
on one positive sample (the RE $E_{r}$ and the image regions $V$
in the image that is paired with it) and $(b-1)$ negative mappings
$\overline{A_{r,s}^{*}}=\left(\overline{a_{r,s,i,j}^{*}}\right)\in\mathbb{R}^{m_{r}\times N}$
where $1\leq s\leq(b-1)$ from negative samples (the RE $E_{r}$ and
negative image regions $V_{s}^{\prime}=\{v_{s,j}^{\prime}\}$ from
images that are not paired with it). We then compute linguistic-induced
visual representations $v_{i}^{*}\in\mathbb{R}^{d}$ and $\overline{v_{s,i}^{*}}\in\mathbb{R}^{d}$
over regions for each word $w_{i}$:

\begin{alignat}{1}
v_{i}^{*} & ={\textstyle \sum_{v_{j}\in V}}\text{norm}_{j}\left(a_{r,i,j}^{*}\right)W_{v}^{\top}v_{j},\\
\overline{v_{s,i}^{*}} & ={\textstyle \sum_{v_{s,j}^{\prime}\in V_{s}^{\prime}}}\text{norm}_{j}\left(\overline{a_{r,s,i,j}^{*}}\right)W_{v^{\prime}}^{\top}v_{s,j}^{\prime},
\end{alignat}
\\
where ``$\text{norm}_{j}$'' is a column normalization operator;
$W_{v}\in\mathbb{R}^{d\times d}$ and $W_{v^{\prime}}\in\mathbb{R}^{d\times d}$
are learnable parameters. We then push them away from each other by
maximizing the linguistic-vision InfoNCE \citep{oord2018representation}:

\begin{equation}
\mathcal{L}_{r}(\delta)=\mathbb{E}_{\mathcal{B}}\left[\sum_{w_{i}\in E_{r}}\textrm{log}\left(\frac{e^{\left\langle W_{w}^{\top}w_{i},v_{i}^{*}\right\rangle }}{e^{\left\langle W_{w}^{\top}w_{i},v_{i}^{*}\right\rangle }\hspace{-0.25em}+\hspace{-0.25em}\sum_{s=1}^{b-1}e^{\left\langle W_{w}^{\top}w_{i},\overline{v_{s,i}^{*}}\right\rangle }}\right)\right].
\end{equation}

This loss maximizes the lower bound of mutual information $\textrm{MI}(V,w_{i})$
between visual regions $V$ and contextualized word embedding $w_{i}$
\citep{gupta2020contrastive}.

Finally, we compute the word-region alignment $A^{*}\in\mathbb{R}^{T\times N}$
by aggregating the RE-image groundings:

\begin{equation}
A^{*}=\frac{1}{|E|}{\textstyle \sum_{r=1}^{|E|}}\tilde{A_{r}^{*}},
\end{equation}
where $\tilde{A_{r}^{*}}\in\mathbb{R}^{T\times N}$ is the zero-padded
matrix of the matrix $A_{r}^{*}$.

Besides making grounding more expressive, this divide-and-conquer
strategy has the extra benefit of augmenting the weak supervising
labels from query-image to RE-image pairs, which increase the amount
of supervising signals (positive pairs) and facilitate better training
of the contrastive learning framework. 

The discovered grounding provides a valuable source of priors for
VQA attention. Existing works \citep{qiao2018exploring,selvaraju2019taking}
use attention priors to regulate the gradient flow of VQA models during
training, hence only constraining the attention weights indirectly.
Unlike these methods, we directly guide the computation of attention
weights via two pathways: through pre-training them without answers,
and by refining in VQA inference on a case-by-case basis.

\subsection{Pre-training VQA Attention \label{sec:Pretraining}}

A typical VQA system seeks to ground linguistic concepts parsed
from the question to the associated visual parts through cross-modal
attention. However, this attention mechanism is guided only indirectly
and distantly through sparse training signal of the answers. This
training signal is too weak to assure that relevant associations can
be discovered. To directly train the attention weights to reflect
these natural associations, we pre-train VQA models by enforcing the
attention weights to be close to the alignment maps $A^{*}$ discovered
through unsupervised grounding in Sec.~\ref{sec:Grounding}.

For joint attention VQA models, this is achieved through minimizing
the Kullback-Leibler divergence between vectorized forms of the VQA
visual attention weights $A$ and the prior grounding scores $A^{*}$:

\begin{align}
\mathcal{L}_{\text{pre-train}} & =KL(\text{norm}\circ\text{vec(}A^{*})\parallel\text{norm}\circ\text{vec}(A)),\label{eq:reconstruction_loss_matrix}
\end{align}
where $\text{norm}\hspace{-0.5mm}\circ\hspace{-0.5mm}\text{vec}$
flattens a matrix into a vector followed by a normalization operator
ensuring such vector sums to one.

For VQA marginalized attention models, we first marginalize $A^{*}=\left(a_{i,j}^{*}\right)\in\mathbb{R}^{T\times N}$
into a vector of visual attention prior:
\begin{equation}
\beta^{*}=\frac{1}{T}{\textstyle \sum_{i=1}^{T}}\text{norm}_{j}(a_{i,j}^{*}).
\end{equation}
The pre-training loss is the KL divergence between the attention weights
and their priors:
\begin{align}
\mathcal{L}_{\text{pre-train}} & =KL(\beta^{*}\parallel\beta).\label{eq:reconstruction_loss}
\end{align}

\subsection{Attention Refinement with Attention Priors \label{sec:refinement}}

\subsubsection{Marginalized attention refinement}

Recall from Sec.~\ref{sec:Preliminary} that a marginalized attention
VQA model computes linguistic attention over $T$ query words $\alpha\in\mathbb{R}^{T}$
and visual attention over $N$ visual regions $\beta\in\mathbb{R}^{N}$.
In this section, we propose to directly refine these attentions using
attention priors $A^{*}=\left(a_{i,j}^{*}\right)\in\mathbb{R}^{T\times N}$
learned in Sec.~\ref{sec:Grounding}. First, $A^{*}$ is marginalized
over rows and columns to obtain a pair of attention priors vectors
$\alpha^{*}\in\mathbb{R}^{T}$ and $\beta^{*}\in\mathbb{R}^{N}$:

\begin{align}
\alpha^{*} & =\frac{1}{N}\sum_{j=1}^{N}\text{norm}_{i}(a_{i,j}^{*});\:\beta^{*}=\frac{1}{T}\sum_{i=1}^{T}\text{norm}_{j}(a_{i,j}^{*}).\label{eq:marginalizing_vis_scores}
\end{align}

We then refine $\alpha$ and $\beta$ inside the reasoning process
through a gating mechanism to return refined attention weights $\alpha^{\prime}$
and $\beta^{\prime}$ in two forms: 

\smallskip{}

\noindent 
\begin{tabularx}{\linewidth}{@{}XX@{}} 

\emph{Additive form:}
\begin{align} 
\label{eqn:additive_combine}
\begin{split}  
\alpha^{\prime} =\lambda\alpha+\left(1-\lambda\right)\alpha^{*},\\ 
\beta^{\prime} =\gamma\beta+\left(1-\gamma\right)\beta^{*}.
\end{split} 
\end{align}

& 
\emph{Multiplicative form:}
\begin{align} 
\label{eqn:mul_combine}
\begin{split}  
\alpha^{\prime} =\textrm{norm}\left(\left(\alpha\right)^{\lambda}\left(\alpha^{*}\right)^{\left(1-\lambda\right)}\right),\\ 
\beta^{\prime} =\textrm{norm}\left(\left(\beta\right)^{\gamma}\left(\beta^{*}\right)^{\left(1-\gamma\right)}\right),
\end{split} 
\end{align}

\end{tabularx}\\
where ``norm'' is a normalization operator; $\lambda\in(0,1)$ and
$\gamma\in(0,1)$ are outputs of learnable gating functions that decide
how much attention priors contribute per words and regions. Intuitively,
these gating mechanisms are a solution to maximizing the agreement
between two sources of information: $\alpha^{\prime}=\text{argmin}\left(\lambda*D(\alpha^{\prime},\alpha)+\left(1-\lambda\right)*D(\alpha^{\prime},\alpha^{*})\right),$
where $D(P_{1},P_{2})$ measures the distance between two probability
distributions $P_{1}$ and $P_{2}$. When $D\equiv$ Euclidean distance,
it gives Eq.~(\ref{eqn:additive_combine}) and when $D\equiv$ KL
divergence between the two distributions, it is Eq.~(\ref{eqn:mul_combine})
\citep{heskes1998selecting} (See the Supplement for detailed proofs).
The same intuition applies for the calculation of $\beta^{\prime}$.

The learnable gates for $\lambda$ and $\gamma$ are implemented as
a neural function $h_{\theta}\left(.\right)$ of visual regions $\bar{v}$
and the question $q$:

\begin{equation}
\lambda=h_{\theta}\left(\overline{v},q\right).\label{eq:lambda_gating_func}
\end{equation}
For simplicity, $\overline{v}$ is the arithmetic mean of regions
in $V$.

For multi-step reasoning, we apply Eqs.~(\ref{eqn:additive_combine},
\ref{eqn:mul_combine}) step-by-step. Since each reasoning step $k$
is driven by an intermediate controlling signal $c_{k}$ (Sec.~\ref{sec:Preliminary}),
we adapt the gating functions to make use of that signal:

\begin{equation}
\lambda_{k}=p_{\theta}\left(c_{k},h_{\theta}\left(\bar{v},q\right)\right).\label{eq:lambda_gating_func_macnet}
\end{equation}

\subsubsection{Joint attention refinement}

In joint attention VQA models, we can directly use matrix $A^{*}=\left(a_{ij}^{*}\right)\in\mathbb{R}^{T\times N}$
without marginalization. With slight abuse of notation, we denote
the output the modulating gate for attention refinement as $\lambda\in(0,1)$
sharing similar role with the gating mechanism in Eq.~(\ref{eq:lambda_gating_func}):

\begin{align}
A^{\prime} & =\begin{cases}
\lambda A+\left(1-\lambda\right)B^{*} & \text{(add.)}\\
\textrm{norm}\left(\left(A\right)^{\lambda}\left(B^{*}\right)^{\left(1-\lambda\right)}\right) & \text{(multi.)}
\end{cases}\label{eq:joint_attn_refine}
\end{align}
where $B^{*}=\text{norm}_{ij}\left(a_{ij}^{*}\right)$.

\subsection{Two-stage Model Training \label{sec:training}}

We perform a two-step pre-training/fine-tuning procedure to train
models using the attention priors: (1) unsupervised pre-training VQA
without answer decoder with attention priors (Sec.~\ref{sec:Pretraining}),
and (2) fine-tune full VQA models with attention refinement using
answers, i.e. by minimizing the VQA loss $-\log P(y\mid q,\mathcal{I})$.

\section{Experiments \label{sec:Experiments}}

\begin{table}[t]
\centering{}\caption{Performance comparison between GAP and other attention regularization
methods using UpDn baseline on VQA v2. Results of other methods are
taken from their respective papers. $^{\dagger}$Our reproduced results.\label{tab:Comparing-att-reg-methods}}
\begin{tabular}{|l|c|c|c|c|}
\hline 
\multirow{2}{*}{{\footnotesize{}Method}} & \multicolumn{4}{c|}{{\footnotesize{}VQA v2 standard val$\uparrow$}}\tabularnewline
\cline{2-5} \cline{3-5} \cline{4-5} \cline{5-5} 
 & {\footnotesize{}All} & {\footnotesize{}Yes/No} & {\footnotesize{}Num} & {\footnotesize{}Other}\tabularnewline
\hline 
{\footnotesize{}UpDn+Attn. Align \citep{selvaraju2019taking}} & {\footnotesize{}63.2} & {\footnotesize{}81.0} & {\footnotesize{}42.6} & {\footnotesize{}55.2}\tabularnewline
{\footnotesize{}UpDn+AdvReg \citep{ramakrishnan2018overcoming}} & {\footnotesize{}62.7} & {\footnotesize{}79.8} & {\footnotesize{}42.3} & {\footnotesize{}55.2}\tabularnewline
{\footnotesize{}UpDn+SCR (w. ext.) \citep{wu2019self}} & {\footnotesize{}62.2} & {\footnotesize{}78.8} & {\footnotesize{}41.6} & {\footnotesize{}54.5}\tabularnewline
{\footnotesize{}UpDn+SCR (w/o ext.) \citep{wu2019self}} & {\footnotesize{}62.3} & {\footnotesize{}77.4} & {\footnotesize{}40.9} & {\footnotesize{}56.5}\tabularnewline
{\footnotesize{}UpDn+DLR \citep{jing2020overcoming}} & {\footnotesize{}58.0} & {\footnotesize{}76.8} & {\footnotesize{}39.3} & {\footnotesize{}48.5}\tabularnewline
{\footnotesize{}UpDn+RUBi$^{\dagger}$ \citep{cadene2019rubi}} & {\footnotesize{}62.7} & {\footnotesize{}79.2} & {\footnotesize{}42.8} & {\footnotesize{}55.5}\tabularnewline
{\footnotesize{}UpDn+HINT \citep{selvaraju2019taking}} & {\footnotesize{}63.4} & \textbf{\footnotesize{}81.2} & {\footnotesize{}43.0} & {\footnotesize{}55.5}\tabularnewline
\hline 
{\footnotesize{}UpDn+GAP} & \textbf{\footnotesize{}64.3} & \textbf{\footnotesize{}81.2} & \textbf{\footnotesize{}44.1} & \textbf{\footnotesize{}56.9}\tabularnewline
\hline 
\end{tabular}\vspace{-2em}
\end{table}

We evaluate our approach (GAP) on two representative marginalized
VQA models: Bottom-Up Top-Down Attention (UpDn) \citep{anderson2018bottom}
for single-shot, MACNet \citep{hudson2018compositional} for multi-step
compositional attention models; and a joint attention model of BAN
\citep{kim2018bilinear}. Experiments are on two datasets: VQA v2
\citep{goyal2017making} and GQA \citep{hudson2019gqa}. Unless stated
otherwise, we choose the additive gating (Eq.~(\ref{eqn:additive_combine})
) for experiments with UpDn and MACNet, and multiplicative forms (Eq.~(\ref{eqn:mul_combine}))
for BAN. Implementation details and extra results are provided in
the supplementary materials.

\subsection{Experimental Results}

\begin{figure*}[t!]

\noindent\begin{minipage}[b]{1\columnwidth}%
\begin{minipage}[c]{0.44\columnwidth}%
\begin{figure}[H]
\begin{centering}
\includegraphics[width=1.03\columnwidth]{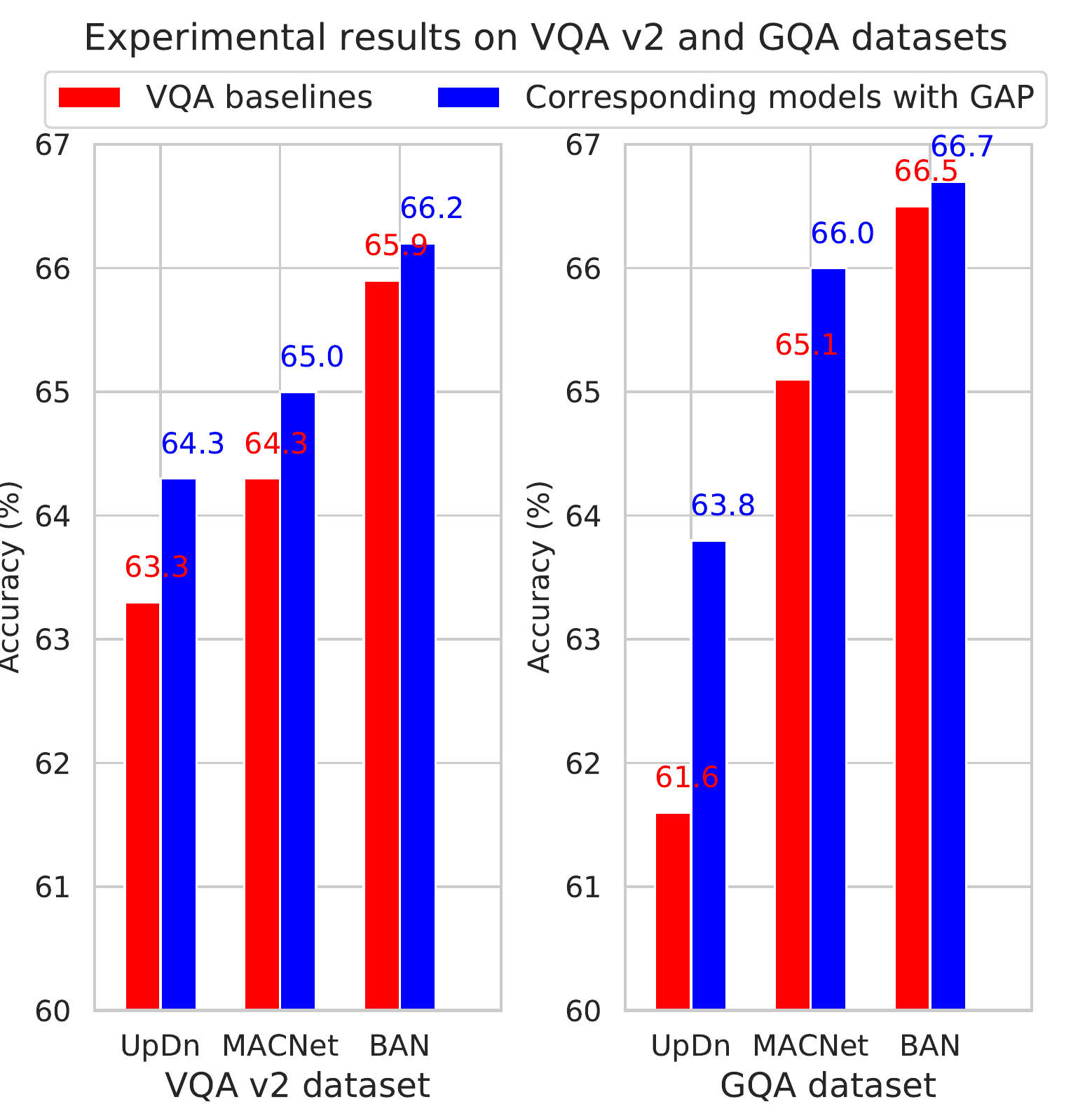}\vspace{1em}
\par\end{centering}
\caption{GAP's universality across different baselines and datasets.\label{fig:comparing_against_baselines_gqa_vqav2}}
\end{figure}
\end{minipage}\hspace{1.5em}%
\begin{minipage}[c]{0.52\columnwidth}%
\begin{figure}[H]
\begin{centering}
\includegraphics[width=0.9\columnwidth]{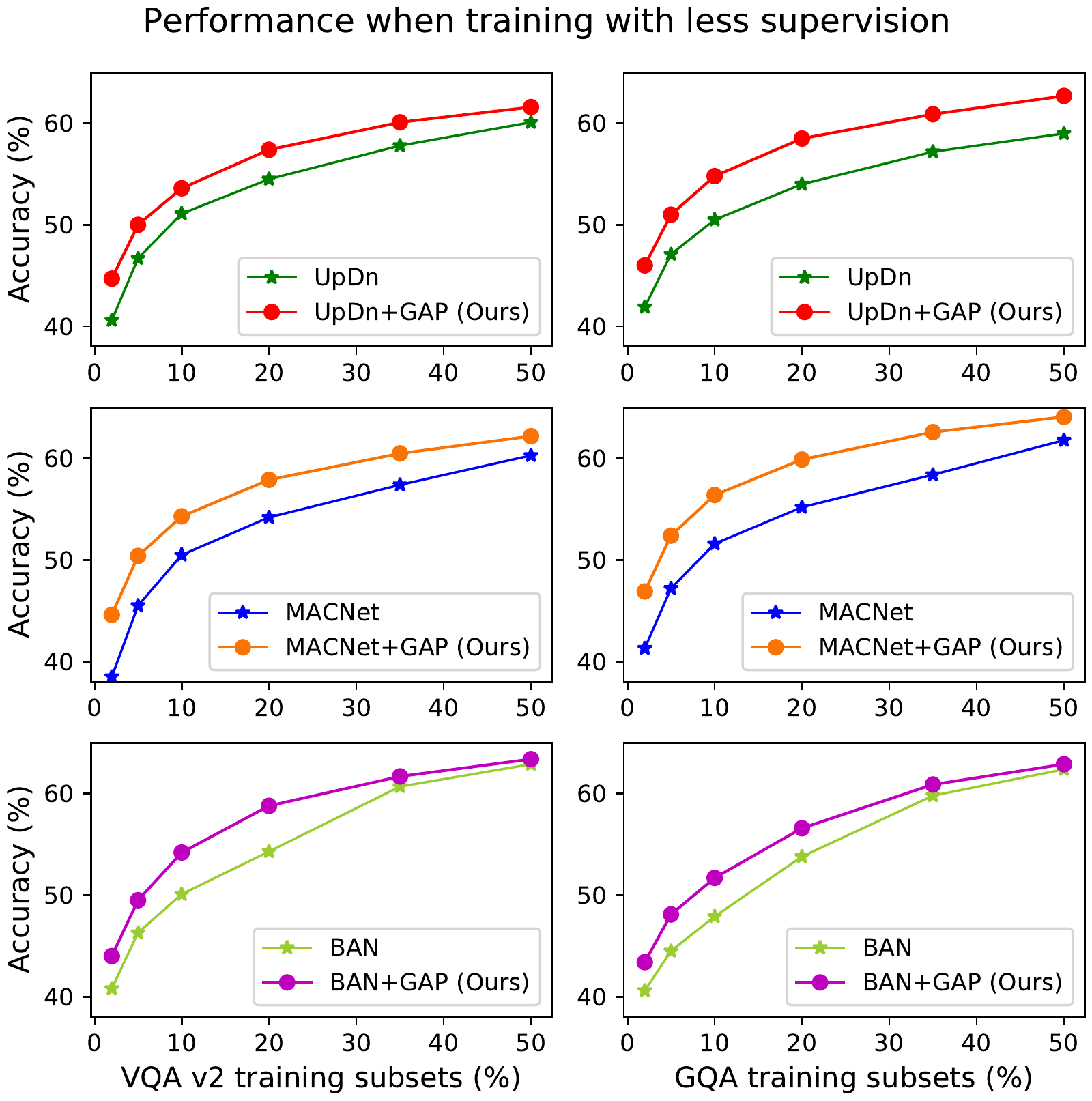}\medskip{}
\par\end{centering}
\caption{GAP improves generalization capability with limited access to grouthtruth
answers.\label{fig:ban_less_supervision}}
\end{figure}
\end{minipage}%
\end{minipage}

\end{figure*}

\paragraph{Enhancing VQA performance }

We compare GAP against the VQA models based on the UpDn baseline that
utilize external priors and human annotation on VQA v2. Some of these
methods use internal regularization: \emph{adversarial regularization}
(AdvReg){\small{} }\citep{ramakrishnan2018overcoming}, \emph{attention
alignment} (Attn. Align) \citep{selvaraju2019taking}; and some use
human attention as external supervision: \emph{self-critical reasoning}
(SCR) \citep{wu2019self} and \emph{HINT} \citep{selvaraju2019taking}.
While these methods mainly aim at designing regularization schemes
to exploit the underlying data generation process of VQA-CP datasets
\citet{agrawal2018don} where it deliberately builds the train and
test splits with different answer distributions. This potentially
leads to overfitting to the particular test splits and accuracy gains
do not correlate to the improvements of actual grounding \citep{shrestha2020negative}.
On the contrary, GAP does not rely on those regularization schemes
but aims at directly improving the learning of attention inside any
attention-based VQA models to facilitate reasoning process. In other
words, GAP complements the effects of the aforementioned methods on
VQA-CP (See the supplemental materials).

Table \ref{tab:Comparing-att-reg-methods} shows that our approach
\emph{(UpDn+GAP)} clearly has advantages over other methods in improving
the UpDn baseline. The favorable performance is consistent across
all question types, especially on ``Other'' question type, which
is the most important and challenging for open-ended answers \citep{teney2021unshuffling,teney2020value}. 

Compared to methods using external attention annotations (\emph{UpDn+SCR},
\emph{UpDn+HINT}), the results suggest that GAP is an effective way
to use of attention priors (in both learning and inference), especially
when our attention priors are extracted in an unsupervised manner
without the need for human annotation.

\paragraph{Universality across VQA models}

GAP is theoretically applicable to any attention-based VQA models.
We evaluate the universality of GAP by trialing it on a wider range
of baseline models and datasets. Fig. \ref{fig:comparing_against_baselines_gqa_vqav2}
summarizes the effects of GAP on UpDn, MACNet and BAN on the large-scale
datasets VQA v2 and GQA. 

It is clear that GAP consistently improves upon all baselines over
all datasets. GAP is beneficial not only for the simple model UpDn,
but also for the multi-step model (MACNet) on which the strongest
effect is when applied only in early reasoning steps where the attention
weights are still far from convergence. 

Between datasets, the improvement is stronger on GQA than on VQA v2,
which is explained by the fact that GQA has a large portion of compositional
questions which our unsupervised grounding learning can benefit from. 

The improvements are less significant with BAN which already has large
capacity model at the cost of data hunger and computational expensiveness.
In the next section, we show that GAP significantly reduces the amount
of supervision needed for these models compared to the baseline.

\begin{table}[t]
\centering{}\caption{Grounding performance of the unsupervised RE-image grounding when
evaluated on out-of-distribution image-caption Flickr30K Entities
test set. \textbf{R}ecall@$k$: fraction of phrases with bounding
boxes that have IOU$\protect\geq$0.5 with top-$k$ predictions.\label{tab:flickr30k_performance}}
\begin{tabular}{|l|c|c|c|c|}
\hline 
{\footnotesize{}Model} & {\footnotesize{}R@1} & {\footnotesize{}R@5} & {\footnotesize{}R@10} & {\footnotesize{}Acc.}\tabularnewline
\hline 
\hline 
{\footnotesize{}$\enskip$Unsupervised RE-image grounding} & {\footnotesize{}14.1} & {\footnotesize{}35.6} & {\footnotesize{}45.5} & {\footnotesize{}45.4}\tabularnewline
\hline 
{\footnotesize{}$\enskip$Unsupervised grounding w/o REs} & \multirow{1}{*}{{\footnotesize{}12.0}} & {\footnotesize{}33.0} & {\footnotesize{}42.9} & {\footnotesize{}44.3}\tabularnewline
\hline 
{\footnotesize{}$\enskip$Random alignment score (10 runs)} & {\footnotesize{}6.6} & {\footnotesize{}28.4} & {\footnotesize{}43.3} & {\footnotesize{}40.7}\tabularnewline
\hline 
\end{tabular}
\end{table}

\paragraph{Sample efficient generalization}

We examine the generalization of the baselines and our proposed methods
when analyzing sample efficiency with respect to the number of annotated
answers required. Fig. \ref{fig:ban_less_supervision} shows the
performance of the chosen baselines on the validation sets of VQA
v2 (left column) and GQA dataset (right column) when given different
fractions of the training data. In particular, when reducing the
number of training instances with groundtruth answers to under 50\%
of the training set, GAP considerably outperforms all the baseline
models in accuracy across all datasets by large margins. For example,
when given only 10\% of the training data, GAP performs better than
the strongest baseline BAN among the chosen ones by over 4.1 points
on VQA v2 (54.2\% vs. 50.1\%) and nearly 4.0 points on GQA (51.7\%
vs. 47.9\%). The benefits of GAP are even more significant for MACNet
baseline which easily got off the track in the early steps without
large data. The results strongly demonstrate the benefits of GAP in
reducing the reliance on supervised data of VQA models. 

\subsection{Model Analysis\label{subsec:Model-Analysis}}

\paragraph{Performance of unsupervised phrase-image grounding }

To analyze the unsupervised grounding aspect of our model, (Sec.~\ref{sec:Grounding}),
we test the grounding model trained with VQA v2 on a mock test set
from caption-image pairs on Flickr30K Entities. This out-of-distribution
evaluation setting will show whether our unsupervised grounding framework
can learn meaningful linguistic-visual alignments.

The performance of our new unsupervised linguistic-visual alignments
using the query grammatical structure is shown in the top row of Table~\ref{tab:flickr30k_performance}.
This is compared against the alignment scores produced by the same
framework but without breaking the query into REs (Middle row) and
the random alignments (Bottom row). There is a 5 points gain compared
to the random scores and over 1 point better than the question-image
pairs without phrases, indicating our linguistic-visual alignments
is a reliable inductive prior for attention in VQA. 

\begin{table*}
\begin{quotation}
\noindent\begin{minipage}[t]{1\columnwidth}%
\hspace{-1.5em}%
\begin{minipage}[t]{0.4\columnwidth}%
\begin{flushleft}
\begin{table}[H]
\centering{}\caption{VQA performance on VQA v2 validation split with different sources
of attention priors.\label{tab:grounding_scores_comparisons}}
\begin{tabular}{|c|>{\raggedright}p{4cm}|c|}
\hline 
\multirow{1}{*}{No.} & \multirow{1}{4cm}{Models} & \multirow{1}{*}{Acc.}\tabularnewline
\hline 
\hline 
1 & $\enskip$UpDn baseline & 63.3\tabularnewline
\hline 
2 & $\enskip$+GAP w/ uniform-values vector & 63.7\tabularnewline
3 & $\enskip$+GAP w/ random-values vector & 63.6\tabularnewline
4 & $\enskip$+GAP w/ supervised grounding & 64.0\tabularnewline
\hline 
5 & $\enskip$+GAP w/ unsup. visual grounding & 64.3\tabularnewline
\hline 
\end{tabular}\vspace{0.47cm}
\end{table}
\par\end{flushleft}%
\end{minipage}\hspace{0.8em}%
\begin{minipage}[t]{0.58\columnwidth}%
\begin{flushright}
\begin{table}[H]
\centering{}\caption{Ablation studies with UpDn on VQA v2.\label{tab:Ablation-studies}}
{\small{}}%
\begin{tabular}{|l|c|}
\hline 
Models & \multirow{1}{*}{Acc.}\tabularnewline
\hline 
\hline 
1.$\enskip$UpDn baseline, $\beta^{\prime}\equiv\beta$ ($\gamma(\theta)\equiv1.0$) & 63.3\tabularnewline
\hline 
\textbf{Attention as priors} & \tabularnewline
2.$\enskip$w/ $\beta^{\prime}\equiv\text{\ensuremath{\beta^{*}}}$($\gamma(\theta)\equiv0.0$) & 60.0\tabularnewline
\textbf{Effects of the direct use of attention priors} & \tabularnewline
3.$\enskip$+GAP w/o 1st stage fine-tuning & 63.9\tabularnewline
4.$\enskip$w/ 1st stage fine-tuning with attention priors & 64.0\tabularnewline
\textbf{Effects of the gating mechanisms} & \tabularnewline
5.$\enskip$+GAP, fixed $\gamma(\theta)\equiv0.5$ & 64.0\tabularnewline
6.$\enskip$+GAP (multiplicative gating) & 64.1\tabularnewline
\textbf{Effects of using visual-phrase associations} & \tabularnewline
7.$\enskip$+GAP (w/o extracted phrases from questions) & 63.9\tabularnewline
\hline 
8.$\enskip$+GAP (full model) & 64.3\tabularnewline
\hline 
\end{tabular}
\end{table}
\hspace{-1.2em}
\par\end{flushright}%
\end{minipage}\vspace{-1.4cm}
\end{minipage}
\end{quotation}
\end{table*}

\paragraph{Effectiveness of unsupervised linguistic-visual alignments for VQA}

We examine the effectiveness of our attention prior by comparing it
with different ways of generating values for visual attention prior
$\beta^{*}$ on VQA performance.  They include: (1) UpDn baseline
(no use of attention prior) (2) uniform-values vector and (3) random-values
vector (normalized normal distribution), (4) supervised grounding
(pretrained MAttNet \citep{yu2018mattnet} on RefCOCO \citep{kazemzadeh2014referitgame}),
and (5) GAP. Table \ref{tab:grounding_scores_comparisons} shows results
on UpDn baseline. GAP is significantly better than the baseline and
other attention priors (2-3-4). Especially our unsupervised grounding
gives better VQA performance than the supervised one (Row 5). This
surprising result suggests that pre-trained supervised model could
not generalize out of distribution, and is worse than underlying grounding
phrase-image pairs extracted unsupervisedly.

\paragraph{Ablation studies}

To provide more insights into our method, we conduct extensive ablation
studies on the VQA v2 dataset (see Table \ref{tab:Ablation-studies}).
Throughout these experiments, we examine the role of each component
toward the optimal performance of the full model. Experiments (1,
2) in Table \ref{tab:Ablation-studies} show that UpDn model does
not perform well with either only its own attention or with the attention
prior itself. This supports our intuition that they complement each
other toward optimal reasoning. Rows 5,6 show that a soft combination
of the two terms is necessary.

Row 7 justifies the use of structured grounding. It shows that phrase-image
grounding gives better performance than question-image pairs only.
In particular, the extracted RE-image pairs improves performance
from 63.9\% to 64.3\%. This clearly demonstrates the significance
of the grammatical structure of questions as an inductive bias for
inter-modality matching which eventually benefits VQA.

\begin{table}[t]
\centering{}\medskip{}
\caption{Grounding scores of UpDn before and applying GAP on GQA validation
split.\label{tab:quantitative_results}}
\begin{tabular}{|l|c|c|c|}
\hline 
{\footnotesize{}Model} & {\footnotesize{}Top-1 attention} & {\footnotesize{}Top-5 attention} & {\footnotesize{}Top-10 attention}\tabularnewline
\hline 
\hline 
{\footnotesize{}UpDn baseline} & {\footnotesize{}14.50} & {\footnotesize{}27.31} & {\footnotesize{}35.35}\tabularnewline
\hline 
{\footnotesize{}UpDn + GAP} & \multirow{1}{*}{{\footnotesize{}16.76}} & {\footnotesize{}29.32} & {\footnotesize{}36.53}\tabularnewline
\hline 
\end{tabular}
\end{table}

\paragraph{Quantitative results}

We quantify the visual attentions of the UpDn model before and after
applying GAP on the GQA validation set. In particular, we use the
grounding score proposed by \citep{hudson2019gqa} to measure the
correctness of the model's attentions weights comparing to the groundtruth
grounding provided. Results are shown in Table \ref{tab:quantitative_results}.
Our method improves the grounding scores of UpDn by 2.26 points (16.76
vs. 14.50) for top-1 attention, 2.01 points (29.32 vs. 27.31) for
top-5 attention and 1.18 points (36.53 vs. 35.35) for top-10 attention.
It is to note that while the grounding scores reported by \citep{hudson2019gqa}
summing over all object regions, we report the grounding scores attributed
by top-$k$ attentions to better emphasize how the attentions shift
towards most relevant objects. This analysis complements the VQA performance
in Table \ref{tab:grounding_scores_comparisons} in a more definitive
confirmation of the role of GAP in improving both reasoning attention
and VQA accuracy.

\begin{figure*}[t]
\begin{centering}
\begin{minipage}[t]{0.92\textwidth}%
\begin{center}
\begin{minipage}[t]{0.48\textwidth}%
\begin{flushleft}
\hspace{-3mm}\includegraphics[width=1.01\columnwidth]{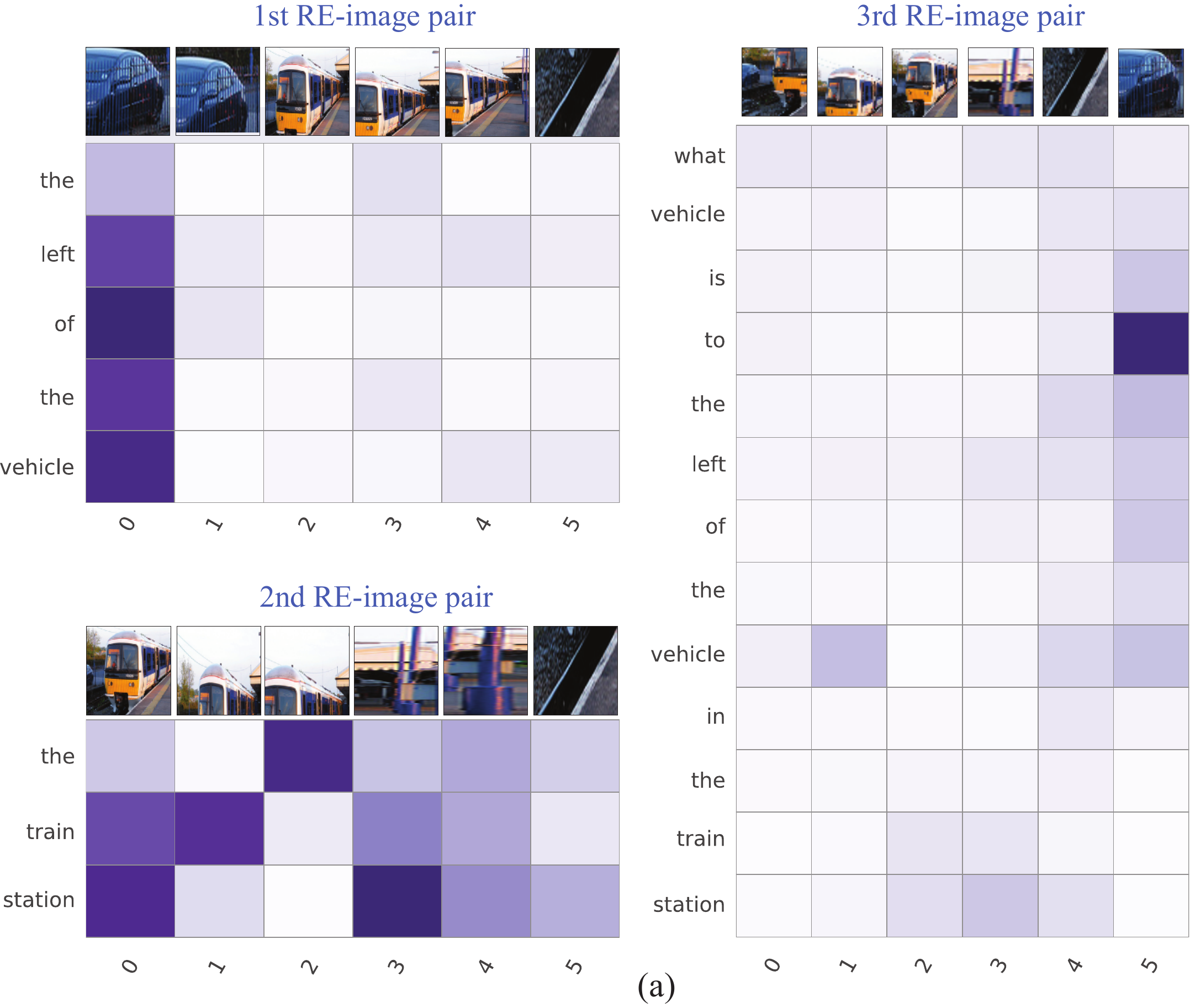}
\par\end{flushleft}%
\end{minipage}\enskip{}{\color{blue}\vrule}\,%
\begin{minipage}[t]{0.48\textwidth}%
\begin{flushright}
\includegraphics[width=1\columnwidth]{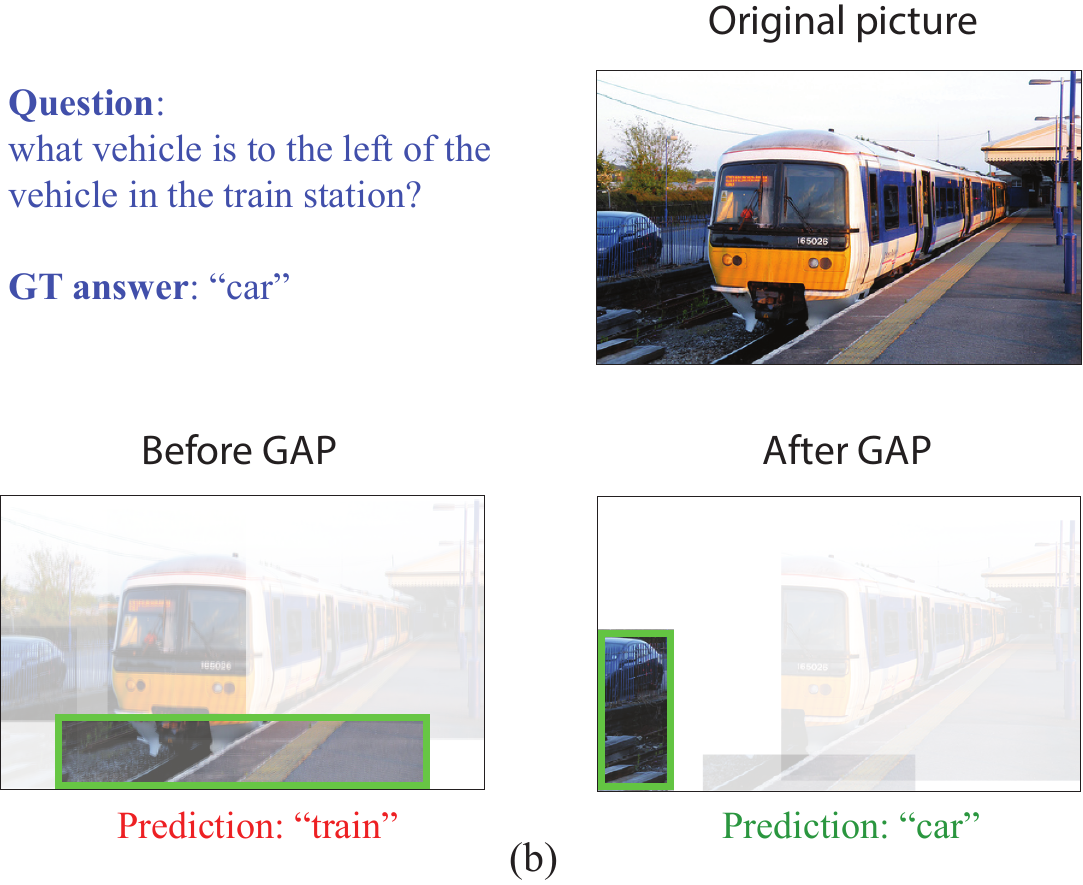}\hspace{-3mm}
\par\end{flushright}%
\end{minipage}
\par\end{center}%
\end{minipage}
\par\end{centering}
\caption{Qualitative analysis of GAP. (a) Region-word alignments of different
RE-image pairs learned by our unsupervised grounding framework. (b)
Visual attentions and prediction of UpDn model before (left) vs. after
applying GAP (right). GAP shifts the model's highest visual attention
(green rectangle) to more appropriate regions while the original puts
attention on irrelevant parts.\label{fig:Qualitative-results}}
\end{figure*}

\paragraph{Qualitative results}

We analyze the internal operation of GAP by visualizing grounding
results on a sample taken from the GQA validation set. The quality
of grounding is demonstrated in Fig. \ref{fig:Qualitative-results}(a)
with the word-region alignments found for several RE-image pairs.
With GAP, these good grounding eventually benefits VQA models by guiding
their visual attentions. Fig. \ref{fig:Qualitative-results}(b) shows
visual attention of the UpDn model before and after applying GAP.
The guided attentions were shifted towards more appropriate visual
regions than attentions by UpDn baseline.

\section{Conclusion \label{sec:Conclusion}}

We have presented a generic methodology to semantically enhance cross-modal
attention in VQA. We extracted the linguistic-vision associations
from query-image pairs and used it to guide VQA models' attention
with Grounding-based Attention Prior (GAP). Through extensive experiments
across large VQA benchmarks, we demonstrated the effectiveness of
our approach in boosting attention-based VQA models' performance and
mitigating their reliance on supervised data. We also showed qualitative
analysis to prove the benefits of leveraging grounding-based attention
priors in improving the interpretability and trustworthiness of attention-based
VQA models. Broadly, the capability to obtain the associations between
words and vision entities in the form of common knowledge is key towards
systematic generalization in joint visual and language reasoning.

{\small{}\bibliographystyle{unsrtnat}
\bibliography{thaole,attn_reg}
\clearpage}{\small\par}

\appendix

\part*{Supplementary material}

In this supplementary material, we provide extra details on the proposed
Grounding as Attention Priors (GAP) method and analysis. They include:
\begin{itemize}
\item Implementation details of the Language and visual embedding (Sec.
\ref{sec:Preliminary} main text)
\item Mathematical proof on the meaning of the attention refinement formulations
(Sec. \ref{sec:refinement} main text).
\item Implementation of the Neural gating functions (Eqs. \ref{eq:lambda_gating_func}
and \ref{eq:lambda_gating_func_macnet} main text).
\item Experiment details including datasets and baselines.
\item Additional experimental results.
\item Additional qualitative analysis.
\end{itemize}

\section{Method details}

\subsection{Language and Visual Embedding}

\paragraph{Textual embedding }

Given a length-$T$ question, we first tokenize it into a sequence
of words and further embed each word into the vector space of 300
dimensions. We initialize the word embeddings with the popular pre-trained
vector representations in GloVe \citep{pennington2014glove}.

To model the sequential nature of the query, we use bidirectional
LSTMs (BiLSTMs) taking as input the word embedding vectors. The BiLSTMs
result in hidden states $\overrightarrow{h_{i}}$ and $\overleftarrow{h_{i}}$
at a time step $i$ for the forward pass and backward pass, respectively.
We further combine every pair $\overrightarrow{h_{i}}$ and $\overleftarrow{h_{i}}$
into a single vector $l_{i}=\left[\overrightarrow{h_{i}};\overleftarrow{h_{i}}\right]$,
where $\left[;\right]$ indicates the vector concatenation operation.
The contextual words are then obtained by gathering these combined
vectors $L=\left\{ l_{i}\mid l_{i}\in\mathbb{R}^{d}\right\} _{i=1}^{T}$.
The global representation $q$ of the query is a combination of the
ends of the LSTM passes $q=\left[\overrightarrow{h_{1}};\overleftarrow{h_{S}}\right]$.

For our grounding framework with contrastive learning, we use a contextualized
word representation extracted by a pre-trained BERT language model
\citep{devlin2018bert} for each word $w_{i}$ in an extracted RE.
These contextualized embeddings are found to be more effective for
phrase grounding \citep{gupta2020contrastive}.

\paragraph{Visual embedding }

Visual regions are extracted by the popular object detection Faster
R-CNN \citep{ren2015faster} pre-trained on Visual Genome \citep{krishna2017visual}.
We use public code\footnote{https://github.com/MILVLG/bottom-up-attention.pytorch}
making use of the Facebook Detectron2 v2.0.1 framework\footnote{https://github.com/facebookresearch/detectron2}
for this purpose. For each image, we extract a set of $N$ RoI pooling
features with bounding boxes $\left\{ \left(a_{j},b_{j}\right)\right\} _{j=1}^{N}$,
where $a_{j}\in\mathbb{R}^{2048},b_{j}\in\mathbb{R}^{4}$ are appearance
features of object regions and bounding box's coordinators, respectively.
We follow \citep{yu2017joint} to encode the bounding box's coordinators
into a spatial vector of 7 dimensions. We further combine the appearance
features with the encoded spatial features by using a sub-network
of two linear transformations to obtain a set of visual objects $V=\left\{ v_{j}\mid v_{j}\in\mathbb{R}^{d^{\prime}}\right\} _{j=1}^{N}$,
where $d^{\prime}$ is the vector length of the joint features of
the appearance features and the spatial features. For ease of reading
and implementation, we choose the linguistic feature size $d$ and
the visual feature size $d^{\prime}$ to be the same.

\subsection{Meaning of Attention Refinement Mechanisms}

In this section, we will give a proof that our choices for attention
refinement in Eqs. (\ref{eqn:additive_combine}, \ref{eqn:mul_combine}
and \ref{eq:joint_attn_refine}) in the main paper are the optimal
solutions with respect to some criteria for probability estimate aggregation. 

Let us consider the generic problem where a system has multiple estimates
$\left\{ P_{i}\left(x\right)\right\} _{i=1}^{n}$ of a true discrete
distribution $P\left(x\right)$ from multiple mechanisms with corresponding
degrees of certainty $\lambda=\left\{ \lambda_{i}\right\} _{i=1}^{n}$.
We first normalize these certainty measures so that they sum to one:
$\lambda_{i}\geq0,\,\sum_{i}\lambda_{i}=1$. We aim at finding a common
distribution $P^{\prime}\left(x\right)$ aggregating the set of distributions
$\left\{ P_{i}\left(x\right)\right\} $ subject to a item-to-set distance
$D$:

\begin{equation}
P^{\prime}\left(x\right)=\underset{P(x)}{\mathrm{argmin}}D_{\lambda}\left(P\left(x\right);\left\{ P_{i}\left(x\right)\right\} _{i=1}^{n}\right).
\end{equation}
This problem can be solved for particular choices of the set-distance
function $D$ that measures the discrepancy between $P\left(x\right)$
and the set $\left\{ P_{i}\left(x\right)\right\} _{i=1}^{n}$ under
the confidence weights $\lambda=\left\{ \lambda_{i}\right\} _{i=1}^{n}$.
We consider several heuristic choices of the function $D$:

\textbf{Additive form:}

If we define the distance to be Euclidean distance from $P\left(x\right)$
to each member distribution $P_{i}\left(x\right)$ of the set, then
the minimized term $D$ becomes

\begin{equation}
D_{\lambda}\left(P\left(x\right);\left\{ P_{i}\left(x\right)\right\} _{i=1}^{n}\right)=\frac{1}{2}\sum_{i}\lambda_{i}\sum_{x}\hspace{-0.5mm}\left(\hspace{-0.5mm}P\left(x\right)\hspace{-0.5mm}-\hspace{-0.5mm}P_{i}\left(x\right)\hspace{-0.5mm}\right)^{2}.
\end{equation}
Here, we minimize $D_{\lambda}$ w.r.t. $P\left(x\right)$:

\[
\partial_{P}D=\sum_{i}\lambda_{i}\left(P\left(x\right)-P_{i}\left(x\right)\right).
\]
By setting this gradient to zero, we have $P^{\prime}\left(x\right)=\sum_{i}\lambda_{i}P_{i}\left(x\right)$.
This explains for the additive form of our attention refinement mechanism
in Eqs. (\ref{eqn:additive_combine} and \ref{eq:joint_attn_refine}
(upper part)) in the main paper where we seek a solution that best
agrees with the grounding prior and the model-induced probability
. \\

\textbf{Multiplicative form:}

If we define $D$ as the weighted sum of the KL divergences between
$P\left(x\right)$ to each member distribution $P_{i}\left(x\right)$
of the set:
\begin{align}
D_{\lambda}\left(P\left(x\right);\left\{ P_{i}\left(x\right)\right\} _{i=1}^{n}\right) & =\sum_{i}\lambda_{i}\textrm{KL}\left(P\left(x\right)\parallel P_{i}\left(x\right)\right)\\
 & =\sum_{i}\lambda_{i}\sum_{x}P\left(x\right)\textrm{log}\frac{P\left(x\right)}{P_{i}\left(x\right)}.
\end{align}
Here, we minimize a Lagrangian with the multiplier $\eta$ of $D_{\lambda}$
w.r.t. $P\left(x\right)$, $L=D+\eta\left(\sum_{x}P\left(x\right)-1\right):$\vspace{-0.3em}

\begin{align}
\partial_{P}L & =\sum_{i}\lambda_{i}\left(\text{log}\frac{P\left(x\right)}{P_{i}\left(x\right)}+1\right)+\eta\\
 & =\text{log}P\left(x\right)-\sum_{i}\lambda_{i}\text{log}P_{i}\left(x\right)+1+\eta\\
 & =\text{log}P\left(x\right)-\text{log}\prod_{i}P_{i}\left(x\right)^{\lambda_{i}}+1+\eta.
\end{align}
Setting this gradient to zero leads to:
\begin{equation}
P^{\prime}\left(x\right)=C\prod_{i}P_{i}\left(x\right)^{\lambda_{i}},
\end{equation}
where $C$ is a calculable constant to normalize $P^{\prime}(x)$
such that its components sum to one. This explains for the multiplicative
form of our attention refinement mechanism in Eqs. (\ref{eqn:mul_combine}
and \ref{eq:joint_attn_refine} (lower part)) in the main paper.

\subsection{Neural Gating Functions}

Here we provide the details of implementation choices for the neural
gating functions in Eqs. (\ref{eq:lambda_gating_func} and \ref{eq:lambda_gating_func_macnet}).
In particular, we use element-wise product between embedded representations
of the input $\overline{v}\in\mathbb{R}^{d}$ and $q\in\mathbb{R}^{d}$
as following:\vspace{-0.3em}

\begin{align}
\lambda & =h_{\theta}\left(\overline{v},q\right)\\
 & =\sigma\left(w_{\lambda}^{\top}\left(\text{ELU}\left(\left(W_{v}^{\top}\overline{v}+b_{v}\right)\varodot\left(W_{q}^{\top}q+b_{q}\right)\right)\right)\right),
\end{align}
where $w_{\lambda}\in\mathbb{R}^{d},\,W_{v}\in\mathbb{R}^{d\times d},\,W_{q}\in\mathbb{R}^{d\times d}$
are learnable weights, $b_{v},b_{q}$ are biases, $\sigma$ is the
sigmoid function and $\varodot$ denotes the Hadamard product. ELU
\citep{clevert2015fast} is a non-linear activation function.

For multi-step reasoning, we additionally takes as input the intermediate
controlling signal $c_{k}\in\mathbb{R}^{d}$ at reasoning step $k$.
Output of the modulating gate in Eq. \ref{eq:lambda_gating_func_macnet}
in the main paper is given by
\begin{equation}
\lambda_{k}=p_{\theta}\left(c_{k},h_{\theta}\left(\overline{v},q\right)\right),
\end{equation}
where\vspace{-0.3em}
\begin{align}
h_{\theta}\left(\overline{v},q\right) & =W_{h}^{\top}\left(\text{ELU}\left(\left(W_{v}^{\top}\overline{v}+b_{v}\right)\varodot\left(W_{q}^{\top}q+b_{q}\right)\right)\right),\\
p_{\theta} & =\sigma\left(w_{\lambda}^{\top}\left(\text{ELU}\left(c_{k}\varodot h_{\theta}\left(\overline{v},q\right)\right)\right)\right).
\end{align}
Here $W_{h}\in\mathbb{R}^{d\times d},\,W_{v}\in\mathbb{R}^{d\times d},\,W_{q}\in\mathbb{R}^{d\times d},\,w_{\lambda}\in\mathbb{R}^{d}$
are learnable weights, and $b_{v},b_{q}$ are biases.

\section{Experiment details}

\subsection{Datasets}

\paragraph{VQA v2}

is a large scale VQA dataset entirely based on human annotation and
is the most popular benchmark for VQA models. It contains 1.1M questions
with more than 11M answers annotated from over 200K MSCOCO images
\citep{lin2014microsoft}, of which 443,757 questions, 214,354 questions
and 447,793 questions in train, val and test split, respectively. 

We choose correct answers in the training split appearing more than
8 times, similar to prior works \citep{teney2018tips,anderson2018bottom}.
We report performance as accuracy calculated by standard VQA accuracy
metric: $\text{min}(\frac{\text{\#humans that provided that answer}}{3},1)$
\citep{antol2015vqa}.

\paragraph*{GQA}

is currently the largest VQA dataset. The dataset contains over 22M
question-answer pairs and over 113K images covering various reasoning
skills and requiring multi-step inference, hence significantly reducing
biases as in previous VQA datasets. Each question is generated based
on an associated scene graph and pre-defined structural patterns.
GQA has served as a standard benchmark for most advanced compositional
visual reasoning models \citep{hudson2019gqa,hu2019language,hudson2019learning,shevchenko2020visual}.
We use the balanced splits of the dataset in our experiments.

\subsection{Baseline Models}

\paragraph{Bottom-Up Top-Down Attention (UpDn)}

UpDn is the first to introduce the use of bottom-up attention mechanism
to $\Problem$ by utilizing image region features extracted by Faster
R-CNN \citep{ren2015faster} pre-trained on Visual Genome dataset
\citet{krishna2017visual}. A top-down attention network driven by
the question representation is used to summarize the image region
features to retrieve relevant information that can be decoded into
an answer. The UpDn model won the VQA Challenge in 2017 and became
a standard baseline $\Problem$ model since then.

\paragraph{MACNet}

MACNet is a multi-step co-attention based model to perform sequential
reasoning where they use $\Problem$ as a testbed. Given a set of
contextual word embeddings and a set of visual region features, at
each time step, an MAC cell learns the interactions between the two
sets with the consideration of their past interaction at previous
time steps through a memory. In particular, an MAC cell uses a controller
to first compute a controlling signal by summarizing the contextual
embeddings of the query words using an attention mechanism. The controlling
signal is then coupled with the memory state of the previous reasoning
step to drive the computation of the intermediate visual attention
scores. At the end of a reasoning step, the retrieved visual feature
is finally used to update the memory state of the reasoning process.
The process is repeated over multiple steps, resembling the way humans
reason over a compositional query. In our experiments, we use a Pytorch
equivalent implementation\footnote{https://github.com/tohinz/pytorch-mac-network}
of MACNet instead of using the original Tensorflow-based implementation.
We choose the number of reasoning steps to be $6$ in all experiments.
For experiments with GAP, we only refine the attention weights inside
the controller (linguistic attention) and the read module (visual
attention) at the first reasoning step where the grounding prior shows
its best effect in accelerating the learning of attention weights,
hence leads to the best performance overall.

\paragraph{Bilinear Attention Networks (BAN)}

BAN is one of the most advanced VQA models based on low-rank bilinear
pooling. Given two input channels (language and vision in the VQA
setting), BAN uses low-rank bilinear pooling to extract the pair-wise
cross interactions between the elements of the inputs. It then produces
an attention map to selectively attend to the pairs that are most
relevant to the answer. BAN also takes advantage of a multimodal residual
networks to improve its performance by repeatedly refining the retrieved
information over multiple attention maps. We use its official implementation\footnote{https://github.com/jnhwkim/ban-vqa}
in our experiments. In order to make the best judgments of the model's
performance with our attention refinement with grounding priors, we
remove the plug-and-play counting module \citep{zhang2018learning}
in the original implementation.

Regarding the choice of hyper-parameters, all the experiments regardless
of the baselines are with $d$=512. The number of visual objects $N$
for each image is $100$ and the maximum number of words $T$ in a
query is set to be the length of the longest query in the respective
dataset. We train all the models using Adam optimizer with a batch
size of $64$. The learning rate is initialized at $10^{-4}$ and
scheduled with the warm up strategy, similar to prior words in VQA
\citep{jiang2018pythia}. Reported results are at the epoch that gives
the best accuracy on the validation sets after $25$ training epochs.

\subsection{Additional Experimental Results}

Apart from the experimental results in Sec. \ref{sec:Experiments}
in the main paper, we provide additional results on VQA-CP2 dataset
\citep{agrawal2018don} to support our claim that GAP complements
related works with regularization schemes. We choose RUBi \citep{cadene2019rubi}
as a representative bias reduction method for VQA with a general linguistic
debiasing technique and yet effective on VQA-CP2 dataset. Table \ref{tab:VQACP-performance}
presents our experimental results with UpDn baseline. As clearly seen,
even though linguistic biases are not the main target, GAP still shows
consistent improvements on top of both UpDn baseline and UpDn+RUBi
baseline. It is to emphasize that applying the regularization by RUBi
for linguistic bias considerably hurts the performance on VQA v2 even
though RUBi largely improves performance on VQA-CP2 test split. GAP
brings the benefits of pre-computed attention priors and rejects the
damage caused by the regularization effects by RUBi to maintain its
excellent performance on VQA v2 while slightly improving the baseline's
performance on VQA-CP2. Looking more closely at the results per question
type on VQA-CP2 (Row 1 vs. Row 2, and Row 3 vs. Row 4), GAP shows
its universal effect on all question types with the strongest effect
on ``Other'' question type which contains open-ended arbitrary questions.
On the other hand, RUBi (Row 3 vs. Row 1) shows its significant impact
only on binary questions ``Yes/No'' but considerably hurts ``Number''
and especially ``Other'' question types. This reveals that the regularization
scheme in RUBi is overfitted to ``Yes/No'' questions specifically
due to the limitation of the data generation process behind this dataset. 

The analysis in this Section is consistent with our results in Figure
4 in the main paper and is clearly evident to GAP's universal effects
in improving VQA performance. The additional results with RUBi shown
in this Section also state GAP's complementary benefits upon the use
of the learning regularization methods targeting only a specific type
of data as such in VQA-CP2.

\begin{table}[t]
\caption{Performance on VQA v2 val split and VQA-CP2 test split with UpDn baseline.\label{tab:VQACP-performance}}

\centering{}%
\begin{tabular}{|c|c|c|c|c|c|c|c|c|}
\hline 
\multirow{2}{*}{{\small{}Model}} & \multicolumn{4}{c|}{{\small{}VQA-CP2 test}} & \multicolumn{4}{c|}{{\small{}VQA v2 val}}\tabularnewline
\cline{2-9} \cline{3-9} \cline{4-9} \cline{5-9} \cline{6-9} \cline{7-9} \cline{8-9} \cline{9-9} 
 & {\small{}Overall} & {\small{}Yes/No} & {\small{}Number} & {\small{}Other} & {\small{}Overall} & {\small{}Yes/No} & {\small{}Number} & {\small{}Other}\tabularnewline
\hline 
\hline 
{\small{}UpDn baseline} & {\small{}40.6} & {\small{}41.2} & {\small{}13.0} & {\small{}48.1} & {\small{}63.3} & {\small{}79.7} & {\small{}42.8} & {\small{}56.4}\tabularnewline
{\small{}UpDn+GAP} & {\small{}40.8} & {\small{}41.2} & {\small{}13.2} & {\small{}48.3} & {\small{}64.3} & {\small{}81.2} & {\small{}44.1} & {\small{}56.9}\tabularnewline
\hline 
{\small{}UpDn+RUBi} & {\small{}48.6} & {\small{}72.1} & {\small{}12.6} & {\small{}46.1} & {\small{}62.7} & {\small{}79.2} & {\small{}42.8} & {\small{}55.5}\tabularnewline
{\small{}UpDn+RUBi+GAP} & {\small{}48.9} & {\small{}72.2} & {\small{}12.8} & {\small{}46.4} & {\small{}64.2} & {\small{}81.4} & {\small{}44.3} & {\small{}56.3}\tabularnewline
\hline 
\end{tabular}
\end{table}

\subsection{Additional Qualitative Analysis}

\begin{figure*}[p]
\begin{centering}
\begin{minipage}[t]{0.92\textwidth}%
\begin{center}
\hspace{-5mm}%
\begin{minipage}[t]{0.47\textwidth}%
\begin{flushleft}
\hspace{-3mm}\includegraphics[width=1.02\columnwidth]{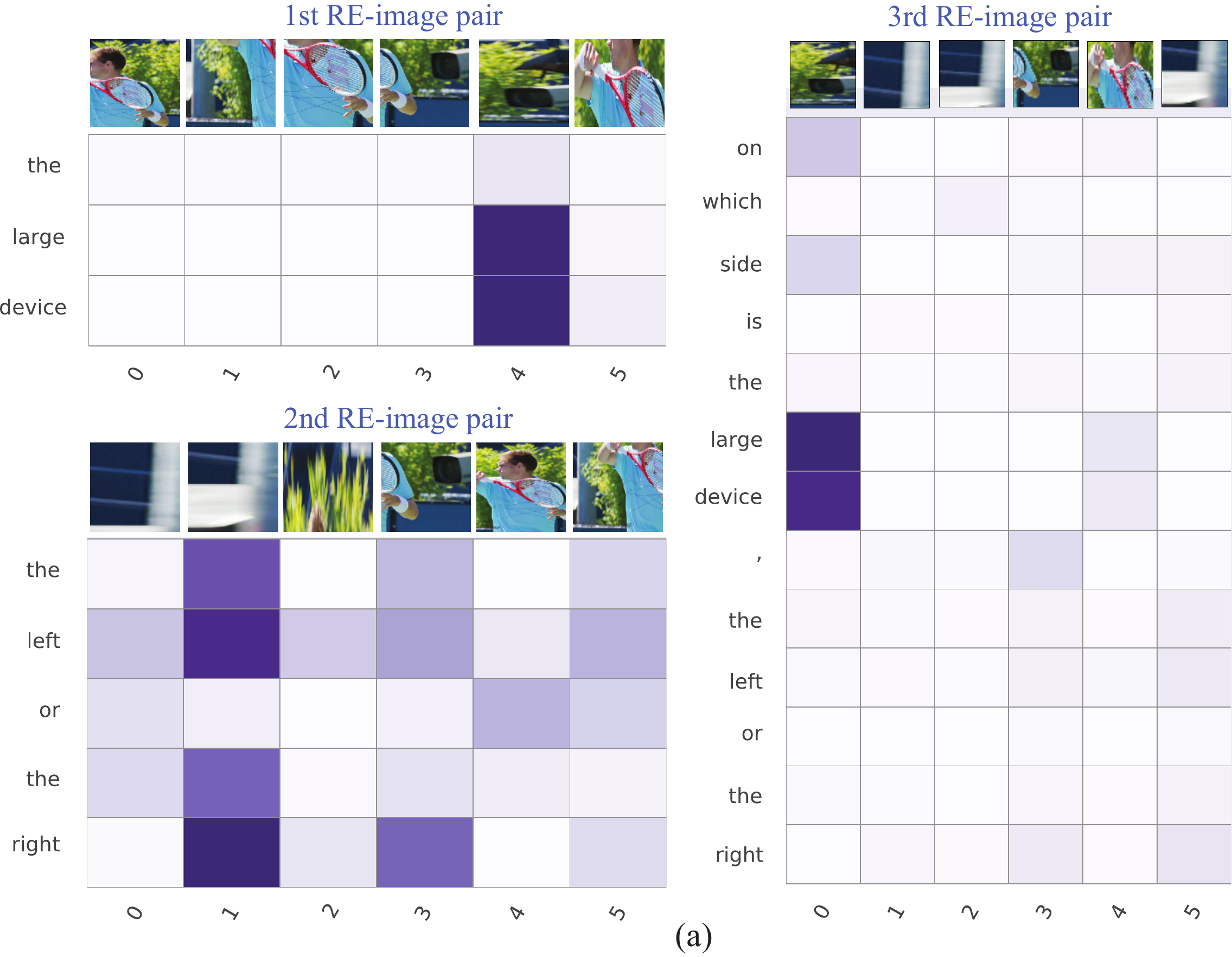}
\par\end{flushleft}%
\end{minipage}\enskip{}{\color{blue}\vrule}\,%
\noindent\begin{minipage}[t]{0.41\textwidth}%
\begin{flushright}
\includegraphics[width=0.99\columnwidth]{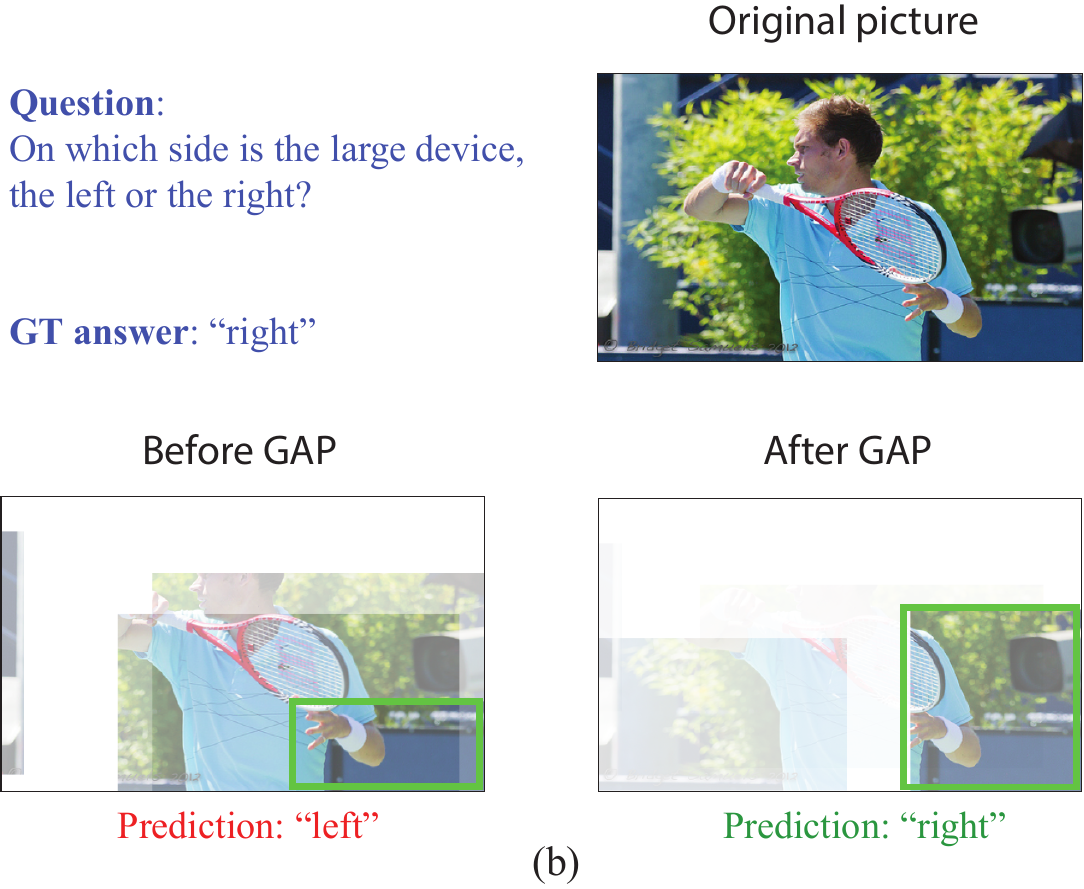}\hspace{-3mm}
\par\end{flushright}%
\end{minipage}\hspace{-5mm}
\par\end{center}%
\end{minipage}
\par\end{centering}
\caption{Qualitative analysis of GAP with UpDn baseline. (a) Region-word alignments
of different RE-image pairs learned by our unsupervised grounding
framework. (b) Visual attentions and prediction of UpDn model before
(left) vs. after applying GAP (right). GAP shifts the model's highest
visual attention (green rectangle) to more appropriate regions while
the original puts attention on irrelevant parts.\label{fig:Qualitative-results-updn}}
\end{figure*}
\begin{figure*}[p]
\begin{centering}
\begin{minipage}[t]{0.92\textwidth}%
\begin{center}
\hspace{-5mm}%
\begin{minipage}[t]{0.47\textwidth}%
\begin{flushleft}
\hspace{-3mm}\includegraphics[width=1.02\columnwidth]{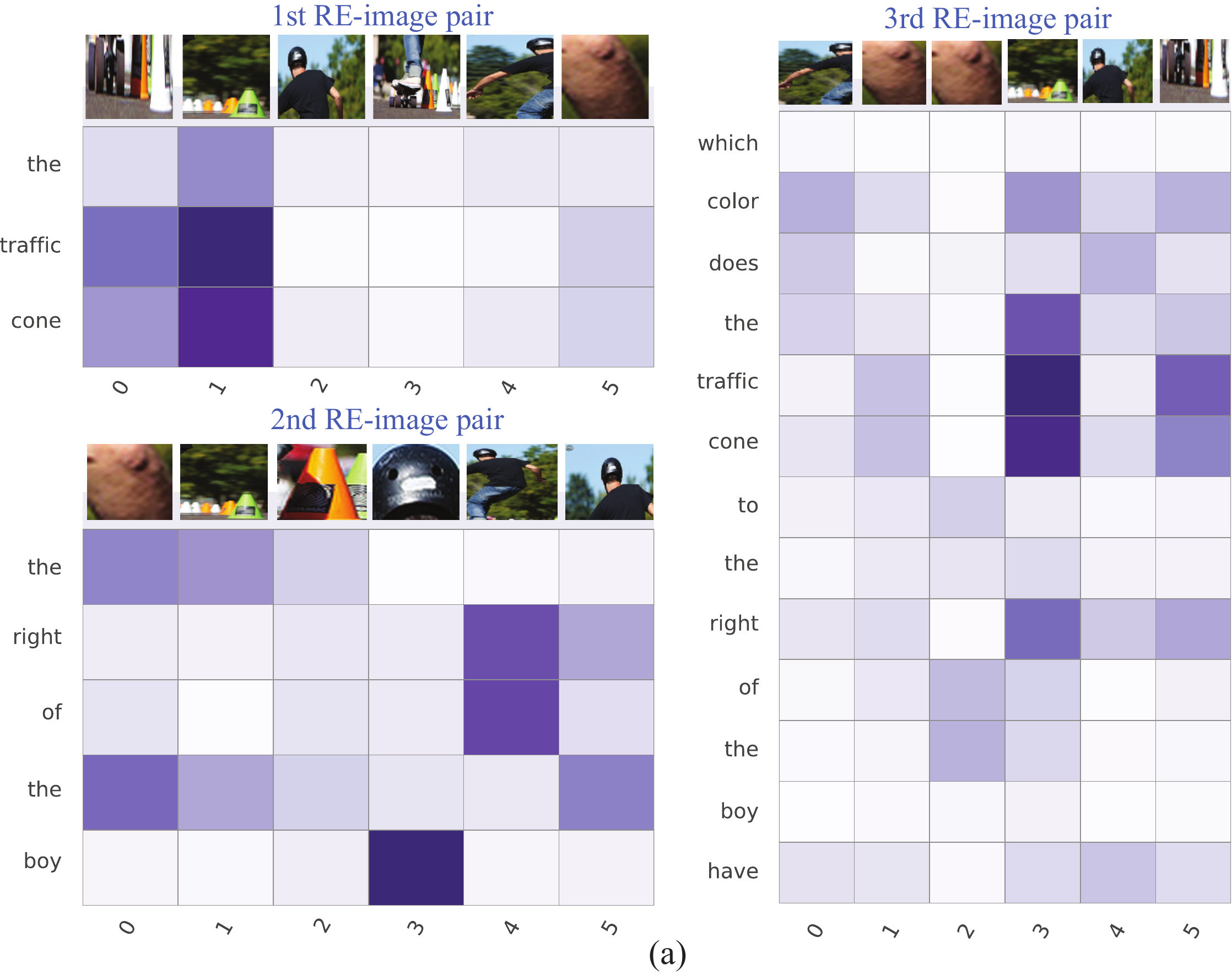}
\par\end{flushleft}%
\end{minipage}\enskip{}{\color{blue}\vrule}\,%
\noindent\begin{minipage}[t]{0.41\textwidth}%
\begin{flushright}
\includegraphics[width=0.85\columnwidth]{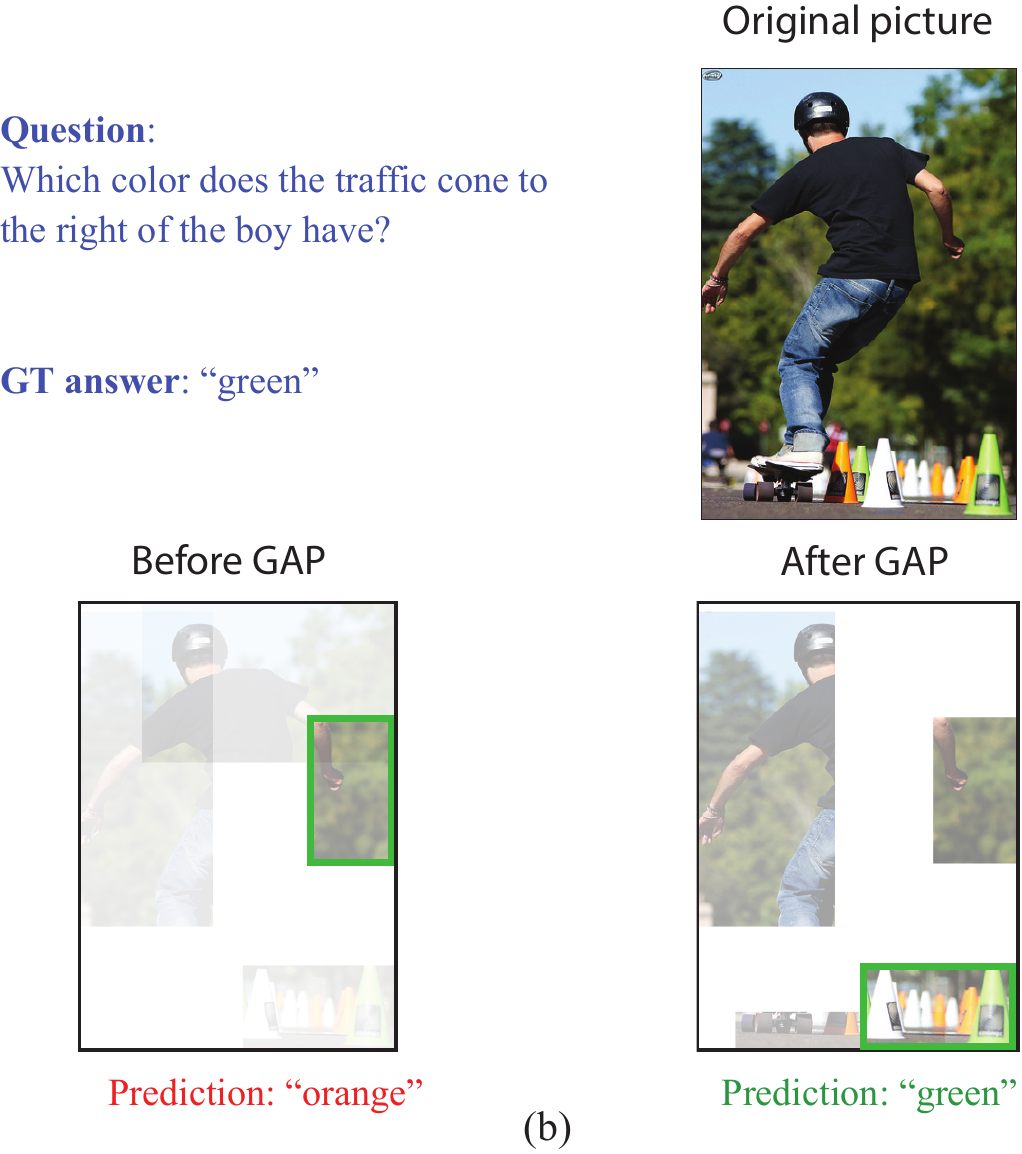}\hspace{-3mm}
\par\end{flushright}%
\end{minipage}\hspace{-5mm}
\par\end{center}%
\end{minipage}
\par\end{centering}
\caption{Qualitative analysis of GAP with MACNet baseline. (a) Region-word
alignments of different RE-image pairs learned by our unsupervised
grounding framework. (b) Visual attentions and prediction of MACNet
model before (left) vs. after applying GAP (right). Visualized attention
weights are obtained at the last reasoning step of MACNet.\label{fig:Qualitative-results-macnet}}
\end{figure*}

Fig.6 in the main paper provides one case of visualization on the
internal operation of our proposed method GAP as well as its effect
on VQA models. We provide more examples here for UpDn baseline (Fig.
\ref{fig:Qualitative-results-updn}) and MACNet baseline (Fig. \ref{fig:Qualitative-results-macnet})
with the same convention and legends. 

In each figure, left subfigures present the linguistic-visual alignments
learned by our unsupervised grounding framework. Right subfigures
compare the visual attentions before and after applying GAP. In all
cases across two different baselines (UpDn and MACNet), GAP clearly
helps direct the models to pay attention to more appropriate visual
regions, partly explaining their answer predictions.

\end{document}